\documentclass[letterpaper,12pt]{article}
\usepackage{authblk}
\usepackage{tabularx} 
\usepackage{amsmath}  
\usepackage{mathtools}
\usepackage{booktabs}
\usepackage{amssymb}
\usepackage{amsthm}
\usepackage{float}
\usepackage{subfig}
\usepackage{algorithm}
\usepackage{algorithmic}
\usepackage{graphicx} 
\usepackage[margin=1in,letterpaper]{geometry} 
\usepackage[final]{hyperref} 
\hypersetup{
	colorlinks=true,       
	linkcolor=blue,        
	citecolor=blue,        
	filecolor=magenta,     
	urlcolor=blue         
}
\usepackage{blindtext}
\usepackage{apacite}

\newcommand*{\email}[1]{%
    \normalsize\href{mailto:#1}{#1}\par
}

\newtheorem{theorem}{Theorem}
\newtheorem{lemma}{Lemma}
\newtheorem{corollary}{Corollary}

\newtheorem{assumption}{Assumption}

\begin{document}

\title{MNL-Bandits under Inventory and Limited Switches Constraints}
\author{Hongbin Zhang$^{1,2}$, Yu Yang$^2$, Feng Wu$^1$ and Qixin Zhang$^2$}
\affil{
$^1$School of Management, Xi'an Jiaotong University \\
$^2$School of Data Science, City University of Hong Kong \\
\email{zhbxjtu@stu.xjtu.edu.cn, yuyang@cityu.edu.hk, fengwu830@126.com, qxzhang4-c@my.cityu.edu.hk}
}
\date{ }
\maketitle

\begin{abstract}
Optimizing the assortment of products to display to customers is a key to increasing revenue for both offline and online retailers. To trade-off between exploring customers' preference and exploiting customers' choices learned from data, in this paper, by adopting the Multi-Nomial Logit (MNL) choice model to capture customers' choices over products, we study the problem of optimizing assortments over a planning horizon $T$ for maximizing the profit of the retailer. To make the problem setting more practical, we consider both the inventory constraint and the limited switches constraint, where the retailer cannot use up the resource inventory before time $T$ and is forbidden to switch the assortment shown to customers too many times. Such a setting suits the case when an online retailer wants to dynamically optimize the assortment selection for a population of customers. We develop an efficient UCB-like algorithm to optimize the assortments while learning customers' choices from data. We prove that our algorithm can achieve a sub-linear regret bound $\tilde{O}\left(T^{1-\alpha/2}\right)$ if $O(T^\alpha)$ switches are allowed. 
Extensive numerical experiments show that our algorithm outperforms baselines and the gap between our algorithm's performance and the theoretical upper bound is small.
\end{abstract}

\section{Introduction}
Assortment selection, which aims at picking a specific set of products to show to customers, is a crucial task in the operational management of retailers. Different assortments may affect customers' purchasing probabilities differently and thus, result in very different revenues for the retailer. To capture customers' choices when facing different assortments, {\it Multinomial Logit (MNL) choice model} is often applied in economics and operations research studies~\cite{talluri2004revenue, caro2007dynamic, kallus2020dynamic}.

In reality, we usually do not know the underlying choice model and we have to learn it from customers' behaviors while optimizing the assortments shown to customers. Therefore, it is nature to cast assortment selection as an online learning problem where the key is to trade off between exploration and exploitation. In addition, considering constraints for online assortment selection can make the problem setting more practical to suit for fruitful applications. In this paper, we study the online assortment selection problem under two important constraints, namely the inventory constraint and the limited switches constraint. 

The inventory constraint specifies that the retailer has a fixed initial inventory of $K$ resources for composing the products, and when a product is sold, a certain amount of resources is consumed. The resources cannot be replenished during the sales horizon. Once one of the resources is depleted, the retailer is forced to stop the sales.

The limited switches constraint forbids the retailer to switch among assortments too many times in the planned sales horizon $T$. Such a constraint is practical for both offline retailers and online retailers. For offline retailers, assortments are shown in displaying shelves and it is very costly to frequently re-arrange the shelves. For online retailers, limiting the switches can be regarded as assigning the same assortment to a group of customers. Such operation is easy to parallelize, which enables us to learn users’ preferences much faster~\cite{perchet2016batched, gao2019batched}. 

To the best of our knowledge, no previous studies have tackled the limited switches constraint for dynamic assortment selection. The switch constraint has already been investigated for Multi-armed Bandits~\cite{simchi2019phase}, where the number of maximum switches $L$ is assumed to be polynomial to the number of arms and all the arms need to be explored. However, when it comes to MNL bandits, if we treat each assortment as an arm, the number of arms is exponential to the number of items $N$. It is not practical to have a planning horizon $T$ exponential to $N$, let alone the maximum number of switches. In addition, when considering the limited switches constraints, the analysis of resource consumption becomes more difficult as we have more constraints when planning the assortments. On the other hand, the inventory constraint also restricts us from using the same assortment for too many times (since we want to reduce assortment switches). In sum, considering both the two realistic constraints, inventory constraint and limited switches constraint, is a challenging task for dynamic assortment selection.


In this paper, we formulate a challenging online assortment selection problem by considering both the inventory constraint and the limited switches constraint. We adopt the Multinomial Logit (MNL) choice model to depict customers' choices and we assume that we do not know the parameters of the underlying MNL model. We develop a UCB-like policy to optimize assortment selection in $T$ planned sales periods while learning the MNL model. We demonstrate that with high probability our policy satisfies both the inventory constraint and the limited switches constraint. We also prove that our policy can achieve a sub-linear regret bound. 

\subsection{Literature Review}
There are two lines of related work—assortment planning and bandits. We will provide a brief review of both fields.

\textbf{Offline assortment optimization.} If the consumer preferences (MNL parameters in our setting) are known a priori, then the related optimal assortment problem we refer to as static assortment optimization problem is well studied. For unconstrained assortment planning problem under the MNL model,  Talluri and van Ryzin \citeyear{talluri2004revenue} and Gallego et al.~\citeyear{gallego2004managing} prove that the optimal assortment belongs to revenue-ordered assortments. 
This important structural result enables the number of candidate assortments from $2^{N}$ to $N$ and will also be used in our policy development. More recent works~\cite{davis2013assortment,desir2014near} consider assortment planning problems under MNL model with various constraints. 



\textbf{Online assortment optimization.} Motivated by large-scale online retailing, relaxing the assumption on prior knowledge of customers’ choice behavior becomes a real need. Caro and Gallien~\citeyear{caro2007dynamic} consider the setting under which demand for products is independent of each other. Rusmevichientong et al.~\citeyear{rusmevichientong2010dynamic} and Saur{\'e} and Zeevi \citeyear{saure2013optimal} consider the problem of minimizing regret under the MNL choice model and present an “explore-first-and-exploit-later” approach. Agrawal et al.\citeyear{agrawal2017thompson,agrawal2019mnl} gave an efficient algorithm that simultaneously explores and exploits in the form of epoch.\citeauthor{kallus2020dynamic}~\citeyear{kallus2020dynamic} considered the problem of dynamic assortment personalization with large, heterogeneous populations and wide arrays of products, and demonstrate the importance of structural priors for effective, efficient large-scale personalization. Closest to our work are~\cite{cheung2017assortment} and~\cite{aznag2021mnl}, where the resources constraints are considered for online assortment selection. As we also consider the limited switches constraint, \cite{cheung2017assortment, aznag2021mnl} cannot solve our problem.

\textbf{Budgeted bandits.} Another closely related stream of literature is the budgeted bandits problem. This is an extension of the MAB problem where pulling an arm generates a revenue and consumes some resources, each with an initial inventory. In the case of random costs, this problem is referred to as the Bandits with Knapsacks (BwK) problem, which was first introduced in \cite{badanidiyuru2013bandits}. Agrawal and Devanur \citeyear{agrawal2019bandits} recently proposed an UCB-based algorithm to achieve a near-optimal regret for a variant of BwK with more general resource constraints. Our work can be interpreted in this framework by considering each assortment as an independent arm. However, directly applying the techniques from \cite{badanidiyuru2013bandits, agrawal2019bandits} would lead to an exponential size arms, which is computationally intractable. Moreover, the regret will be linear to the number of assortments.

\textbf{Combinatorial bandits.} Our problem is also related to the Combinatorial Bandits \cite{kveton2015tight,chen2013combinatorial}. By interpreting each assortment as a superarm, our problem can be transformed into this framework. However, in our setting, the reward of an assortment depends on all products present in the subset, whereas in Combinatorial Bandits, the reward is assumed to be independent. Moreover, the revenue generated by the assortment is not even monotonic, as introducing new products in the assortment may reduce the probability of purchasing the most profitable items and may thus reduce the expected revenue. Hence, we cannot directly use the existing Combinatorial Bandits to obtain a low regret policy. 

\textbf{Limited switches.} Modeling the switch number as a hard constraint is common in the literature. \citeauthor{cheung2017dynamic}\citeyear{cheung2017dynamic} considers a dynamic pricing model where the demand function is unknown beforehand but belongs to a known finite set, and limits the number of price changes. \citeauthor{simchi2019phase}\citeyear{simchi2019phase} considers the stochastic MAB bandit problem under a general switching constraint. Neither of the two papers considers the existence of non-replenishable resource constraints. Some batched MAB problems can also be viewed as a switch limit~\cite{perchet2016batched,gao2019batched}.

\subsection{Our contributions}
\begin{itemize}
    \item We are the first to consider both limited switches and resources constraints, and reveal the relationship between the regret and the number of switches.
    \item We propose a UCB-like policy that incurs a sublinear regret under this setting. Specifically, if we set the warm-start period $\tau=\tilde{O}\left(T^{1 / 2}\right)$ and the number of epochs $q(L,K)=\tilde{O}\left(T^{\alpha}\right)$, then the regret is at most $\tilde{O}\left(T^{1-\frac{\alpha}{2}}\right)$. The regret is sub-linear to the total periods $T$. Our regret bound matches the optimal regret bound $\Omega(T^{1/2})$ with respect to $T$ when $\alpha=1$.
    \item To deal with the issue of solving an exponential size LP similar to \cite{cheung2017assortment}, we formulate a quadratic size LP and prove that it is equivalent to the exponential size LP. Such equivalent formulation not only makes our UCB-like algorithm computationally efficient, but also helps accelerate optimizing assortment under only inventory constraints~\cite{cheung2017assortment}. 
    \item Our UCB-like policy can also be applied to optimize assortments for a population of a fixed size $\lfloor \frac{T-\tau}{q(L,K)}\rfloor$, which is adjustable by modifying the switches limit $L$. 
    \item Extensive numerical experiments show that our policy outperforms the exploration-exploitation (EE) policy and is computationally efficient.
\end{itemize}

\section{Problem Formulation}
The dynamic assortment selection problem with the multinomial logit (MNL) choice model, also called MNL-bandit, is a fundamental problem in online learning and operations research. In this problem, the retailer has a set of resources $\mathcal{K}=\{1, \ldots, K\}$ with $C(k)$ units of initial inventory for each resource $k$, where $C(k)=Tc(k)$ and $T$ is the number of periods. Each product in the set of products $\mathcal{N}=\{1, \ldots, N\}$ is composed by the $K$ resources. The sale of one product $i$ generates a revenue of $r(i)\in [0,1]$, and consumes $a(i,k)\in [0,1]$ units of resource $k$, for each $k\in \mathcal{K}$. Product 0 is the "no-purchase" product and $r(0)=a(0,k)=0$ for all $k\in \mathcal{K}$. In each $t\in [T]$, the following sequence of events occur.

First, a customer arrives in period $t$. Second, the retailer offers an assortment $S_{t}\in \mathcal{S}$ to the customer, where $\mathcal{S}=\{S\subseteq\mathcal{N}\}$ is the collection of all subsets of $\mathcal{N}$. When seeing the assortment $S_t$, the customer purchases an item $I_{t}\in S_{t}\cup \{0\}$. If $I_{t}=0$, it means the customer leaves without buying anything. Then the retailer earns a revenue of $r(I_{t})$, and the corresponding resources are consumed. We then update the resources as $C(k)=C(k)-a(I_{t},k)$ for all $K \in \mathcal{K}$.

Customers' purchasing behaviour is depicted by the MNL choice model. If we know the true preference vector $v^{*}=(v^*_1,\dots,v^*_N)$, where $v^*_i$ denotes the relative utility of purchasing item $i$, the probability of purchasing item $i$ given assortment $S$ is
\begin{equation}\label{equation: MNL_model}
    \varphi(i,S\mid v^{*})=\frac{v_{i}^{*}}{1+\sum_{j \in S}v_{j}^{*}}
\end{equation}
Hence, the probability of a customer purchasing nothing is $\varphi(0,S\mid v^{*})=1/(1+\sum_{j \in S}v_{j}^{*})=1-\sum_{i\in S} \varphi(i,S\mid v^{*})$. For any product $i\in \mathcal{N}\backslash S$, $\varphi(i,S\mid v^{*})=0$.

$\textbf{Limited switches}$. We define the number of switches during the time horizon $T$ as 	$\Psi_{T}= \sum_{t=1}^{T} \mathbb{I}\left[S_{t} \neq S_{t+1}\right]$, and limit the number of switches to be no greater than $L$, that is $\Psi_{T}\leq L$. 

$\textbf{Regret Minimization.}$ The retailer doesn't know the exact value of $v^{*}$ and only knows that $v_{i}^{*}\in [1/R,R]$ for all $i\in \mathcal{N}$. The retailer needs to learn the parameters based on the previous presented assortments $S_{1}, \ldots, S_{t-1}$ as well as the customers’ actual choices $I_1,\ldots,I_{t-1}$, and develop a non-anticipatory policy to determine the assortment $S_{t}$ to offer at time $t$. The policy aims to maximize the total revenue $\sum_{t=1}^{T}r(I_{t})$, subject to the resource and switch constraints. Once one of the resources depleted, the retailer stop the sale. Equivalently, we model this problem as minimizing the regret as follows.
\begin{equation}\label{equation: regret}
    	Reg(T)=T*OPT(LP(v^{*}))-\sum_{t=1}^{T}r(I_{t})
\end{equation}
subject to: $\sum_{t=1}^{T}a(I_{t},k)\leq C(k)$, for all $k \in \mathcal{K}$, and $\Psi_{T}\leq L$. The linear model $LP(v)$ is formulated as follows:
\begin{equation}\label{equation:ub_model}
\begin{aligned}
LP(v)=\max & \sum_{S \in \mathcal{S}} R(S \mid v) y(S) \\
\text { s.t. } & \sum_{S \in \mathcal{S}} A(S, k \mid v) y(S) \leq c(k) \quad \forall k \in \mathcal{K} \\
& \sum_{S \in \mathcal{S}} y(S)=1, \quad y(S) \geq 0 \quad \forall S \in \mathcal{S}
\end{aligned}
\end{equation}
where  $R(S \mid v)=\sum_{i\in S}r(i)\varphi(i,S\mid v) $ is the expected revenue when offering an assortment $S$ under the preference vector $v$. $A(S,k \mid v)=\sum_{i\in S}a(i,k)\varphi(i,S\mid v) $ is the expected consumed amount of resource $k$. $y(S)$ denotes the probability of choosing $S$. $LP(v)$ aims to maximize the expected revenue in a period under the preference vector $v$. 
$T*OPT(LP(v^{*}))$ is the upper bound of the expected revenue generated from the total $T$ periods.

\begin{theorem}\label{theorem:ub}
\cite{badanidiyuru2013bandits} For any non-anticipatory policy $\pi$ that satisfies the resource constraints with probability 1 , the following inequality holds:
\begin{equation}\label{equation:ub_state}
    T*OPT\left(L P\left(v^{*}\right)\right) \geq \mathbb{E}\left[\sum_{t=1}^{T} r\left(I_{t}^{\pi}\right)\right],
\end{equation}
where $I_{t}^{\pi}$ denotes the random product purchased by the period $t$ customer under policy $\pi$.
\end{theorem}

\section{UCB-like Policy}\label{sec:UCB}
Our algorithm is described in Algorithm~\ref{alg:ucb}. For better exposition, we first let periods 1 to $\tau$ be the warm-start period. During this period, we offer single item assortments in order to estimate $v^{*}$. Then, we run our algorithm in an epoch schedule. Specifically, we first compute a key index $q(L,K)$ in line 8 based on the switching budget $L$, the number of products $N$, and the number of resource constraints $K$, which denotes the number of epochs. Then, we divide the rest periods from $\tau+1$ to $T$ into $q(L,K)$ epochs of equal lengths. Each epoch $\ell$ contains periods from $T_{\ell -1}+1$ to $T_{\ell}$. $N_{\ell}(S)$ is the number of times that the assortment $S$ is offered in epoch $\ell$, and $n_{\ell}(S)=\sum_{k=1}^{\ell}N_{k}(S)$ is the number of times that the assortment $S$ is offered before epoch $\ell$.

\begin{algorithm}
	\caption{UCB-like Policy}
	\begin{algorithmic}[1]\label{alg:ucb}			
		\ENSURE $C(k)=Tc(k)$
		\FOR{$i=1,\dots,N$}
		\FOR{$t=(i-1)\tau/N+1$ to $i\tau/N$}
		\STATE Offer $S_{t}=\{i\}$, observe outcome $I_{t}$
		\STATE For all $k \in \mathcal{K}, C(k) \leftarrow C(k)-a\left(I_{t}, k\right)$
		\ENDFOR	
		\ENDFOR
		\STATE Mark the last period as $T_{0}$
		\STATE Calculate $q(L,K)=\lfloor\frac{L-N}{K+1}\rfloor$.
		\begin{equation}\label{eq5}
		    			T_{\ell}=\ell*\lfloor \frac{T-\tau}{q(L,K)}\rfloor+\tau, \quad \forall \ell=1, \ldots, q(L,K);
		\end{equation}
		\FOR{$ \ell=1,2,\dots,q(L,K)$} 
		\STATE Compute the MLE $\hat{v}^{\ell}$ based on ${{(S_{t},I_{t})}}^{T_{\ell-1}}_{t=1}$
		\STATE Solve UCB-LP($\hat{v}^{\ell},n^{\ell-1},\omega$) (model~\eqref{model:Choice_LP}) for an optimal $y_{\ell}$	
		\FOR{$ t=T_{\ell-1}+1,\dots,T_{\ell}$} 
		\STATE Sample $S_{t}$ with probability  $y_{\ell}(S_{t})$ 
		\ENDFOR			
		\FOR {$\forall S \in \mathcal{S}$ and $N_{\ell}(S)>0$}
		\STATE Offer assortment $S$ for $N_{\ell}(S)$ consecutive periods. \\
		\STATE Observe $I_{t}$; For all $k \in \mathcal{K}, C(k) \leftarrow C(k)-a\left(I_{t}, k\right)$\\
		\STATE Stop the algorithm once one of the resources is exhausted.
		\ENDFOR		
		\ENDFOR
	\end{algorithmic}
\end{algorithm}

At the beginning of each epoch $\ell$, we first need to compute the MLE $\hat{v}_{i}^{\ell}$ for each product $i$. We adopt the Maximum Likelihood Estimation to obtain $\hat{v}^{\ell}$. By introducing an auxiliary variable $\theta_i$ for each $v_i$ where $e^{\theta_i}=v_i$, we have the negative log likelihood function as
\begin{equation}\label{equation:negative_log}
	\begin{split}
		\mathcal{L}_{\ell-1}(\theta)&=-\log \left[\prod_{t=1}^{T_{\ell-1}}\left(\frac{v_{I_{t}}}{1+\sum_{i\in S_{t}}v_{i}}\right)\right]\\&=-\sum_{t=1}^{T_{\ell-1}}\log\frac{v_{I_{t}}}{1+\sum_{i\in S_{t}}v_{i}}\\
		&=-\sum_{t=1}^{T_{\ell-1}}\log v_{I_{t}}-\log(1+\sum_{i\in S_{t}}v_{i})\\
		&=-\sum_{t=1}^{T_{\ell-1}}\theta_{I_{t}}-\log(1+\sum_{i\in S_{t}}e^{\theta_i})
	\end{split}
\end{equation}

The policy estimates the parameter $v$ only $q(L,K)$ times instead of $O(T)$ times, which helps relieve the computational burden. To learn the estimation of $v$, we apply the Maximum Likelihood estimation to minimize Eq.~\ref{equation:negative_log}. We have that the Hessian matrix of Eq.~\eqref{equation:negative_log}
\begin{equation}\label{equation:hess_matrix}
		H_{\ell-1}(\theta) \succeq \frac{1}{R(1+NR)^{2}} \times diag(n^{\ell-1}_1,\cdots,n^{\ell-1}_N),
\end{equation}
where $diag(n^{\ell-1}_1,\cdots,n^{\ell-1}_N)$ is a positive definite diagonal matrix. Therefore, Eq.~\eqref{equation:negative_log} is convex and we minimize Eq.~\eqref{equation:negative_log} to obtain $\hat{v}^{\ell}$.

Next, we solve the following UCB-LP$(\hat{v}^{\ell},n^{\ell-1},\omega)$ model for an optimal solution $y_{\ell}$ as follows.

\begin{equation}\label{model:Choice_LP}
	\begin{split}
		\max \sum_{S \subseteq \mathcal{N}}\sum_{i \in S}r(i) \Big( &\frac{\hat{v}_{i}^{\ell}}{1+\sum_{j\in S}\hat{v}_{j}^{\ell}}+\varepsilon (n^{\ell-1}_i) \Big) y_{\ell}(S)\\
		\text { s.t. }  \sum_{S \in \mathcal{N}}\sum_{i \in S}a(i,k) \Big( &\frac{\hat{v}_{i}^{\ell}}{1+\sum_{j\in S}\hat{v}_{j}^{\ell}}-\varepsilon ( n^{\ell-1}_i) \Big) y_{\ell}(S)\\
		&\leq(1-\omega) c(k) \quad \forall k \in \mathcal{K}\\
		\sum_{S \in \mathcal{N}} y_{\ell}(S)=1, & \quad y_{\ell}(S) \geq 0 \quad \forall S \in \mathcal{N}
	\end{split}
\end{equation}
Notice that the reward of each assortment is overestimated and the consumption of each assortment is underestimated. Hence, similar to~\cite{agrawal2019bandits}, we replace $c(k)$ by $(1-\omega)c(k)$ in the above model to tighten the constraints. The parameters in Eq.~\eqref{model:Choice_LP} are defined as follows
\begin{equation}\label{equation:epsilon}
    \varepsilon(n)=\frac{(\sqrt{N}+1)\Psi}{\sqrt{n}}
\end{equation}
\begin{equation}\label{equation:n_i}
    n^{\ell-1}_i=\sum_{t=1}^{T_{\ell-1}} \mathbb{I}\left(i \in S_{t}\right)
\end{equation}
\begin{equation}\label{equation:Psi}
	\Psi=\frac{R(1+N R)^{2}}{2}\sqrt{2+4 \log \frac{2T^{1/2}q(L,K)(K+1)N}{\delta}}
\end{equation}

\begin{equation}\label{equation:omega}
\begin{split}
   	&\omega  =\frac{1}{T\min _{k \in \mathcal{K}} c(k)}\Bigg(4(\sqrt{N}+1)\sqrt{1+\frac{NT}{\tau q(L,K)}}\Psi\sqrt{N^{2}T}\\
	&+\sqrt{2T\log\frac{4(K+1)}{\delta}}+\frac{2N^{2} \Psi}{\sqrt{\tau}}\sqrt{2T\log\frac{4(K+1)}{\delta}}+\tau\Bigg) 
\end{split}
\end{equation}

Before showing how to solve Eq.~\eqref{model:Choice_LP}, we first demonstrate that our Algorithm~\ref{alg:ucb} does not violate the limited switches constraint. Since there are $K+1$ constraints in Eq.~\eqref{model:Choice_LP}, the optimal solution $y_{\ell}$ contains at most $K+1$ non-zero entries, which means at most $K+1$ assortments are with positive probabilities to be selected. During each epoch, the policy chooses at most $K+1$ kinds of assortments, thus making at most $K+1$ switches between them. So there are at most $N+(K+1)*q(L,K)\leq L$ switches, satisfying the definition of the $L$-switch learning policy.
\begin{theorem}[Limited Switches]\label{th:switch}
The number of assortment switches made by Algorithm~\ref{alg:ucb} is at most $L$.
\end{theorem}

\noindent \textbf{Remark} We can view our algorithm as a parallel assortment planning algorithm, where at the beginning of each period $t$ we decide all the assortments for $\lfloor \frac{T-\tau}{q(L,K)}\rfloor$ customers. Therefore, our algorithm can be applied to optimize assortments for a population of customers, where the population size $\lfloor \frac{T-\tau}{q(L,K)}\rfloor$ is adjustable by modifying the switches limit $L$. Existing epoch-based MNL-bandits~\cite{agrawal2017thompson, agrawal2019mnl, aznag2021mnl} cannot deal with this as their epochs do not have a uniform length.


There are $2^n$ decision variables and $K+1$ constraints in the UCB-LP$(\hat{v}^{\ell},n^{\ell-1},\omega)$ model (Eq.~\eqref{model:Choice_LP}). Solving the exponential size LP model always requires using column generation. Due to the existence of $\varepsilon(n^{\ell-1}_i)$, it is difficult to solve the sub-problem of column generation. Hence, we reformulate an equivalent Compact LP as follows.

\begin{equation}\label{model:Compact_LP}
\begin{split}
    \max _{\left(x_{0}, \boldsymbol{x}, \boldsymbol{y}\right) \in \mathbb{R} \times \mathbb{R}_{+}^{n+n^{2}}} &\sum_{i \in N} r_{i}\Big[\left(\hat{v}_{i}^{\ell}+ \varepsilon(n^{\ell-1}_i)
    \right) x_{i}+\varepsilon(n^{\ell-1}_i) \sum_{j \in N} \hat{v}_{j}^{\ell} y_{i j}\Big] \\
    s.t. \quad\quad~~~~ &\sum_{i \in N} a(i,k)\Big[\left(\hat{v}_{i}^{\ell}- \varepsilon(n^{\ell-1}_i)
    \right) x_{i}\Big.\\
    &\Big.-\varepsilon(n^{\ell-1}_i) \sum_{j \in N} \hat{v}_{j}^{\ell} y_{i j}\Big] \leq (1-\omega) c(k) \quad \forall k \in \mathcal{K}\\
   & x_{0}+\sum_{i \in N} \hat{v}_{i}^{\ell} x_{i}=1 \\
    & x_{i} \leq x_{0} \quad \forall i \in N \\
    &y_{i j} \leq \min\{x_{i},x_j\} \quad \forall i, j \in N
\end{split}
\end{equation}
There are only $n^2 +n+1$ decision variables and $2n^2 +n+K+1$ constraints in the Compact LP. Therefore, the Compact LP of quadratic size can be easily solved.

We show that UCB-LP (Eq.~\eqref{model:Choice_LP}) and Compact LP model (Eq.~\eqref{model:Compact_LP}) are equivalent to each other and we can recover an optimal solution to the former by using the latter in a way similar to \cite{cao2020revenue}.\\

\begin{theorem}[Equivalence of LP Formulations]\label{theorem:EquivLP}
 The optimal objective values of the UCB-LP (Eq.~\eqref{model:Choice_LP}) and Compact LP (Eq.~\eqref{model:Compact_LP}) are the same. Furthermore, the optimal values of the dual variables for the first constraint in the UCB-LP and Compact LP are the same.
\end{theorem}

Let $\left(x_{0}^{*}, \boldsymbol{x}^{*}, \boldsymbol{y}^{*}\right)$ be a basic optimal solution to the Compact LP. We index the products so that $x_{1}^{*} \geq x_{2}^{*} \geq \ldots \geq x_{n}^{*}$. Defining the set $S_{i}=\{1, \ldots, i\}$ with $S_{0}=\varnothing$, for each $i=0,1, \ldots, n$, we set
$$
\hat{w}\left(S_{i}\right)=\left(x_{i}^{*}-x_{i+1}^{*}\right)\left(1+V\left(S_{i}\right)\right)
$$
where we follow the convention that $x_{n+1}^{*}=0$. Note that $x_{0}^{*} \geq x_{i}^{*}$ for all $i \in N$ by the third constraint in the Compact LP, we have $\hat{w}\left(S_{0}\right)=x_{0}^{*}-x_{1}^{*} \geq 0$.

\begin{theorem}[Recovering an Optimal Solution]\label{theorem:RecoverOPT}
For a basic optimal solution $\left(x_{0}^{*}, \boldsymbol{x}^{*}, \boldsymbol{y}^{*}\right)$ to the Compact $\mathrm{LP}$, let $\hat{\boldsymbol{w}}=\{\hat{w}(S): S \subseteq N\}$ be constructed as in the Recovery formula with $\hat{w}(S)=0$ for all $S \notin\left\{S_{0}, S_{1}, \ldots, S_{n}\right\}$. Then, $\hat{\boldsymbol{w}}$ is an optimal solution to the UCB-LP model. 
\end{theorem}

We present the regret bound of Algorithm~\ref{alg:ucb}. First, we make an assumption of the inventory of resources that is also adopted in \cite{cheung2017assortment} to ensure that no resource is depleted before period $\tau$.

\begin{assumption}\cite{cheung2017assortment}\label{Assumption:assumption}
The learning phase length $\tau$ satisfies: For all $k \in \mathcal{K}, \tau \sqrt{\log \frac{4 N K}{\delta}} \leq T c(k)$; 
\end{assumption}

\begin{theorem}[Regret bound]\label{theorem:Regbound}
Suppose $\tau$ satisfies Assumption ~\ref{Assumption:assumption}. The proposed UCB-like policy satisfies all resource constraints and incurs a regret at most
\begin{equation}\label{equation:regretBound}
    \begin{split}
        \textup{Bound}=&(1+\frac{1}{\min _{k \in \mathcal{K}} c(k)})\Bigg(4(\sqrt{N}+1) \sqrt{1+\frac{NT}{\tau q(L,K)}}\Psi\sqrt{N^{2}T} \\
        &+\Big( \frac{2N^{2} \Psi}{\sqrt{\tau}}+N+1 \Big) \sqrt{2T\log\frac{4(K+1)}{\delta}}+\tau \Bigg)
    \end{split}
\end{equation}
with probability $1-\delta$.
\end{theorem}

In particular, the regret varies as $q(L,K)$ and $\tau$ vary. The bigger the $q(L,K)$, the smaller the regret. Specifically, if we set $\tau=\tilde{O}\left(T^{1 / 2}\right)$ and $L=O(T^\alpha)$ (which means $q(L,K)=\tilde{\Theta(1/\min c_k)}\left(T^{\alpha}\right)$), then the regret is at most $\tilde{O}\left(T^{1-\frac{\alpha}{2}}\right)$. Note that setting $L=O(T^\alpha)$ for $\alpha \leq 1$ is reasonable in real applications as the number of assortment is exponential to $N$ and a practical $T$ is often in low-degree polynomial to $N$.

When $\alpha=1$, which means $L=T$ and there is actually no limited switches constraint, our regret bound in Theorem~\ref{theorem:Regbound} is $\tilde{O}(T^{1/2})$, which matches the regret lower bound of online assortment selection without any constraints~\cite{agrawal2016near} with respect to $T$. 

\section{Regret Analysis}
We demonstrate the key steps to prove the regret bound in Theorem~\ref{theorem:Regbound}. To bound the regret, we first decompose it into several parts.
\begin{equation}\label{equation:regret_decomposition}
    \begin{aligned}
        & \mathbb{P}\left[\text { Regret } \leq \text { Bound }\right] \\
        =& \mathbb{P}\left[T*OPT\left(LP\left(v^{*}\right)\right)-\sum_{t=1}^{T} r\left(I_{t}\right) \leq \text{Bound}\right] \\
        \geq& \mathbb{P}\left[\left\{T*OPT\left(LP\left(v^{*}\right)\right)-\sum_{t=1}^{T} r\left(I_{t}\right) \leq \text{Bound}\right\}\right.\\
        &\left.\cap\left\{\sum_{t=1}^{T} a\left(I_{t}, k\right) \leq T c(k) \text { for all } k .\right\}\right]\\
        \geq& \mathbb{P}\left[\left\{T*OPT\left(LP\left(v^{*}\right)\right)-\sum_{t=1}^{T} r\left(I_{t}\right) \leq \text{Bound}\mid \mathcal{A}\right\}\right.\\
        &\left.\cap\left\{\sum_{t=1}^{T} a\left(I_{t}, k\right) \leq T c(k) \text { for all } k .\right\}\mid \mathcal{A}\right]\mathbb{P}\left[\mathcal{A}\right]
    \end{aligned}
\end{equation}
The definition of event $\mathcal{A}$ is given blow. Then, under event $\mathcal{A}$, several other lemmas are given to prove the regret. Finally, we prove the resources bound and the regret bound.

\begin{lemma}[Event $\mathcal{A}$]\label{lemma:event A}
 For each $i\in [N], \ell\in [q(L,K)]$, define the following events
$$
	\mathcal{A}_{i,\ell}=\left\{\left|\frac{\partial L_{\ell-1}}{\partial \theta_i}\right|_{\theta=\theta^{*}}  \leq \Delta\right\},
$$
where
$$
\Delta=\sqrt{2 n^{\ell-1}_i \left(1+2 \log \frac{2\sqrt{T}q(L,K)N}{\delta}\right)}.
$$
Let $\mathcal{A}_{\ell}=\bigcap_{i=1}^{N}\mathcal{A}_{i,\ell}$ and $\mathcal{A}=\bigcap_{i=1}^{N}\bigcap_{\ell=1}^{q(L,K)}\mathcal{A}_{i,\ell}$.
Then, for each $i,\ell$, $\mathbb{P}_{\pi}\left(\mathcal{A}_{i, \ell}\right) \geq 1-\delta/2q(L,K)N$, for each $\ell$, $\mathbb{P}_{\pi}\left(\mathcal{A}_{ \ell}\right) \geq 1-\delta/2q(L,K)$.  Moreover, $\mathbb{P}_{\pi}\left(\mathcal{A}\right) \geq 1-\delta/2$.
\end{lemma}

Under the event $\mathcal{A}$, we first bound the estimated parameters and true parameters.
\begin{theorem}[Parameter Bound]\label{theorem:parameter}
Consider the sales process $\left\{S_{t}, I_{t}\right\}_{t=1}^{T_{\ell-1}}$ generated by Algorithm \ref{alg:ucb}. Under event $\mathcal{A}$, the following inequality holds.

\begin{equation}
	\sum_{i=1}^{N}\left(\sqrt{n^{\ell-1}_i} \log \left|\frac{\hat{v}_{i}^{\ell}}{v_{i}^{*}}\right|-\Psi\right)^{2} \leq N \Psi^{2}
\end{equation}
It implies that the confidence bound
\begin{equation}\label{equation:Parameter_bound}
	\left|\log \frac{\hat{v}_{i}^{\ell}}{v_{i}^{*}}\right| \leq \varepsilon(n^{\ell-1}_i)
\end{equation}
where the definition of $\varepsilon(n^{\ell-1}_i)$ is given in equation~\eqref{equation:epsilon}.
\end{theorem}

Then, we establish the Lipschitz continuity of $\varphi(i, S \mid v)$ in $\log v$. 
\begin{lemma}[Lipschitz continuity of $\varphi(i, S \mid v)$]\label{lemma:lipschitz}
For all $v, v^{\prime} \in \mathbb{R}_{>0}^{\mathcal{N}}, b \in(0,1)^{\mathcal{N}}$ and $S \subseteq \mathcal{N}$, the following inequality holds:
\begin{equation}\label{equation:chain_ineq}
    \sum_{i \in S} b(i)\Big[\varphi(i, S \mid v)-\varphi\left(i, S \mid v^{\prime}\right)\Big] \leq \sum_{i \in S}b(i)\left|\log \frac{v_{i}}{v_{i}^{\prime}}\right| .
\end{equation}
\end{lemma}

Combining Lemma ~\ref{lemma:lipschitz} and Theorem ~\ref{theorem:parameter}, we have the following Corollary.
\begin{corollary}\label{corollary:1}
Let $B(S \mid v)=\sum_{i \in S} b(i) \varphi(i, S \mid v)$. Under event $\mathcal{A}$, the event $E_{t}$	such that
\begin{equation}\label{eq22}
\begin{split}
    	B\left(S \mid \hat{v}^{\ell}\right)&-\sum_{i \in S}b(i) \varepsilon(n^{\ell-1}_i) \leq B\left(S \mid v^{*}\right)\\
    	&\leq B\left(S \mid \hat{v}^{l}\right)+\sum_{i \in S}b(i) \varepsilon(n^{\ell-1}_i)
\end{split}
\end{equation}
holds for all $S \in \mathcal{S}$ and $b \in(0,1)^{N}$.
\end{corollary}

We show that if we set $\omega$ properly, Algorithm~\ref{alg:ucb} does not violate the inventory constraint with high probability. This is equivalent to proving that no product is entirely consumed before time $T$ with high probability.

\begin{lemma}[The bound of resources]\label{lemma:ResourceBound}
When we set $\omega$ as Eq.\eqref{equation:omega}, no resources are consumed before T with probability with probability at least $1-K \delta /2(K+1)$, that is.
$$ 
\mathbb{P}\left[\left\{\sum_{t=1}^{T} a\left(I_{t}, k\right) \leq T c(k) \text { for all } k .\right\}\mid \mathcal{A}\right]\ge 1-K \delta /2(K+1).
$$
\end{lemma}

We also establish the relationship of $\operatorname{OPT}\left(\textup{UCB-LP}\left(\hat{v}^{\ell}, n^{\ell-1}, \omega\right)\right)$ and $\operatorname{OPT}\left(L P\left(v^{*}\right)\right)$.
\begin{lemma}\label{lemma:4}
Conditional on the event $E_{t}$, we have
$$
\operatorname{OPT}\left(\textup{UCB-LP}\left(\hat{v}^{\ell}, n^{\ell-1}, \omega\right)\right) \geq(1-\omega) \operatorname{OPT}\left(L P\left(v^{*}\right)\right)
$$
\end{lemma}

Finally, under the event $\mathcal{A}$, we bound the regret.

\begin{lemma}[The bound of regret]\label{lemma:Regretbound}
Suppose $\tau$ satisfies Assumption~\ref{Assumption:assumption}. Under event $\mathcal{A}$, the probability of regret bound is:

$
\mathbb{P}\left[{T*OPT\left(LP\left(v^{*}\right)\right)-\sum_{t=1}^{T} r\left(I_{t}\right) \leq \text{Bound}}\mid \mathcal{A}\right]\ge 1- \delta /2(K+1).
$
\end{lemma}

Combining Lemma~\ref{lemma:event A}, Lemma~\ref{lemma:ResourceBound}, and Lemma~\ref{lemma:Regretbound}, Theorem ~\ref{theorem:Regbound} can be proved.

\section{Numerical Simulations}
Note that by modifying $L$ we can adjust the value of $q(L,K)$ and vice versa. Let $q(L,K)=T^{\alpha}$. We evaluate the performance of our policy under three settings with synthetic data, $\alpha=0$, $\alpha=1/2$ and $\alpha=1$. When $q(L,K)=T^{0}$, it actually means an exploration then exploitation (EE) policy. The setting $q(L,K)=T$ ($\alpha=1$) means we need to estimate parameters and solve a LP model at every step as there is actually no limited switches constraint. Note that setting $\alpha=1$ equals solving the problem without limited switches constraint~\cite{cheung2017assortment} by using our equivalent Compact LP to accelerate the computation.

In our experiments, we consider sales horizon lengths in $\{250, 500, 750, 1000, 1500, 2000, 5000,$ $10000, 20000, 30000, 40000\}$. Let $\Gamma=\left(N,K,R\right)$ represent a combination of model parameters.We consider four combinations of the model parameters 
$\Gamma_{1}=\left( 10,5,3\right), \Gamma_{2}=\left(15,6,5\right),\Gamma_{3}=\left( 25,8,7\right),\Gamma_{4}=\left(50,12,12\right)$.
The tuples $\Gamma_{1},\Gamma_{2},\Gamma_{3}$ and $\Gamma_{4}$ are sorted in ascending order of difficulty. As the number of products increases to 50, the number of assortments increases up to $1.0 \times 10^{15}$, which makes the problem extremely difficult to solve.

For each $\left(\Gamma_{i}, T\right)$, we generated 5 random instances and ran 10 times under each setting. All the instances and code can be found in supplementary materials. For each run on each instance, we calculated the revenue-to-optimum ratio, which is the ratio between the revenue earned by our policy and the theoretical upper bound $T*OPT\left(L P\left(v^{*}\right)\right)$. We then calculated the average revenue-to-optimum ratio over all runs and all instances.

$\varepsilon(n)$ and $\omega$ in model~\eqref{model:Choice_LP} appear as artifacts of the proof. We omitted them in our numerical simulations. Once one of the resources is depleted, we stop running our algorithm. 

\begin{figure}[hbt!]
    \centering
    \includegraphics[width=1\linewidth]{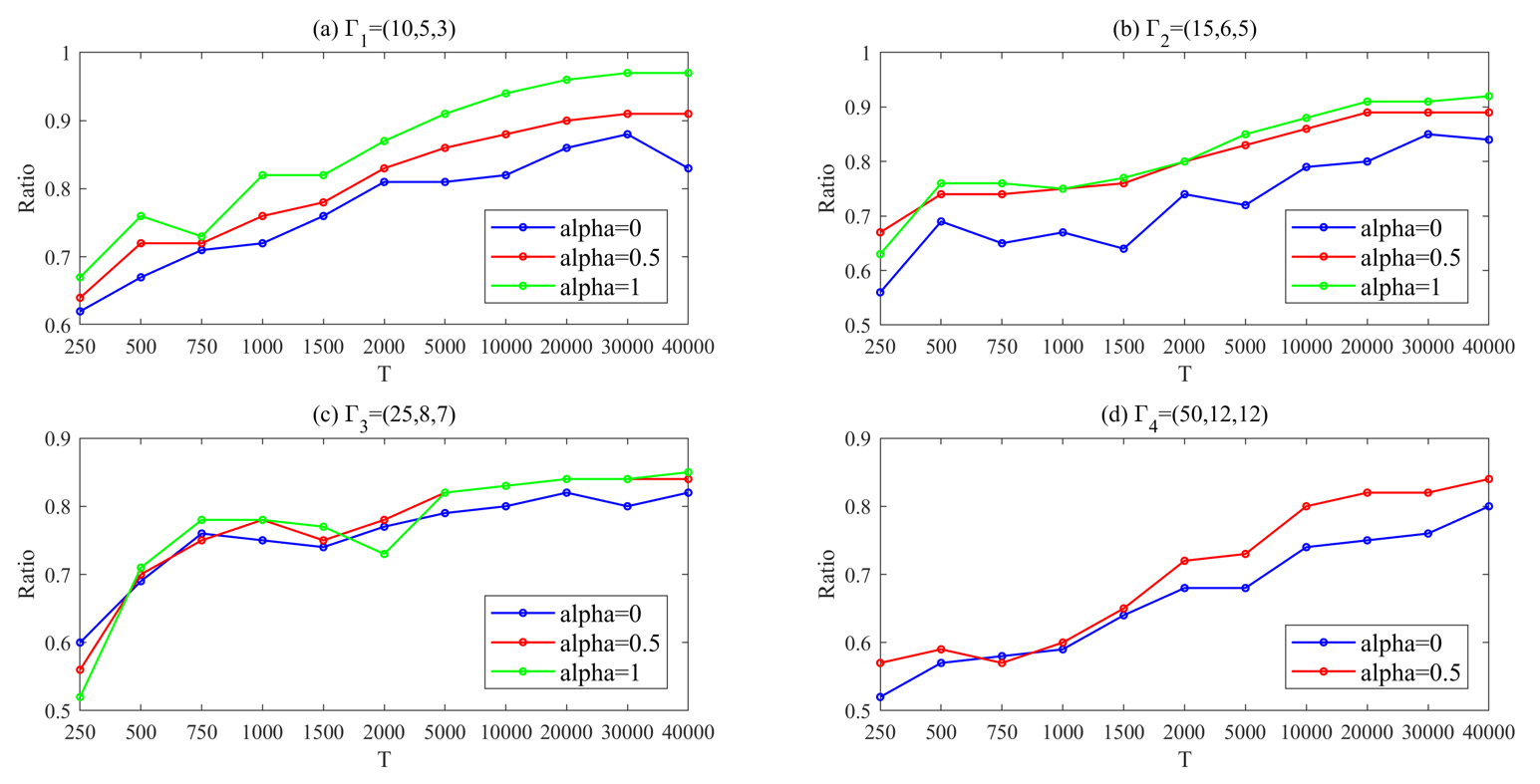}
    \caption{Revenue-to-optimum ratios}\label{fig:r}
\end{figure} 
Fig.1 shows how the average revenue-to-optimum ratio under each $\Gamma_{i}$ increases as $T$ increases. Our policy performs well even for $\Gamma_{4}=\left(50,12,12\right)$ when $\alpha=1/2$, where the revenue-to-optimum ratio can achieve up to 0.84. On the other hand, we notice that the optimal revenue-to-optimum ratio decreases when the number of products in an assortment increases, which verifies that $\Gamma_{1}$, $\Gamma_{2}$, $\Gamma_{3}$ and $\Gamma_{4}$ are sorted in ascending order of difficulty. An important point we want to mention is that setting $\alpha=1/2$ outperforms setting $\alpha=0$ (this setting can be viewed as exploration-then-exploitation policy) under all parameters. In Fig.~\ref{fig:r}.(d), we do not report the performance of $\alpha=1$ since it is too time consuming where the running time of one instance is more than 5 hours when $T>10000$, while it takes only minutes to run for $\alpha=1/2$. 
    \begin{figure}[hbt!]
      \centering
        \subfloat[$T=30000$]{
          \includegraphics[width=0.5\linewidth]{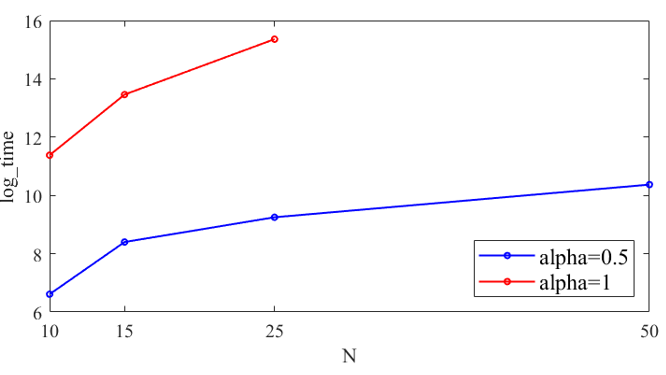}\label{fig:r1}
        }
        \subfloat[$\Gamma_{3}=(25,8,7)$]{
          \includegraphics[width=0.5\linewidth]{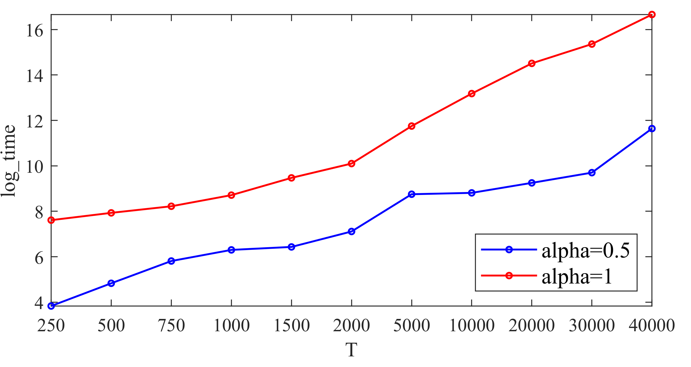}\label{fig:r2}
        }
        \caption{The running time in milliseconds}\label{fig:sca}
    \end{figure} 
    
We compare the scalability of $\alpha=1/2$ (there is a limited switch constraint) and $\alpha=1$ (there is no limited switches constraint) in Fig.~\ref{fig:sca}. We find that not considering the limited switches constraint ($\alpha=1$) is difficult to scale as the number of items increases. For example, when we offer 50 products in an assortment, setting $\alpha=1$ usually takes more than 10 hours for one instance. Moreover, the running time of $\alpha=1/2$ also scales better than the running time of $\alpha=1$ when we vary $T$. This is because when there is a constraint of limited switches, we do not need to do MLE and solving model~\eqref{model:Choice_LP} for every period $t$.

\section{Conclusion}
In this paper, we studied a dynamic assortment optimization problem under an unknown MNL model, and considered limited resources and switches constraints. To maximize the total expected revenue, we designed a UCB-like policy which achieves a sub-linear regret bound that is optimal with respect to $T$. Specifically, if we set the warm-start period $\tau=\tilde{O}\left(T^{1 / 2}\right)$ and the number of epochs $q(L,K)=\tilde{O}\left(T^{\alpha}\right)$, then the regret is at most $\tilde{O}\left(T^{1-\frac{\alpha}{2}}\right)$. To reduce the computational burden caused by solving an exponential size LP, we formulated an equivalent LP of only quadratic size. We conducted extensive numerical experiments. The experimental results show that our policy is efficient and it outperforms the existing exploration-exploitation (EE) policy significantly. For future work, we plan to incorporate both item price and contextual information into the choice model for optimizing assortment selection.

\bibliography{ref}

\begin{thebibliography}{}

\bibitem [\protect \citeauthoryear {%
Abbasi-Yadkori%
, P{\'a}l%
\BCBL {}\ \BBA {} Szepesv{\'a}ri%
}{%
Abbasi-Yadkori%
\ \protect \BOthers {.}}{%
{\protect \APACyear {2011}}%
}]{%
abbasi2011improved}
\APACinsertmetastar {%
abbasi2011improved}%
\begin{APACrefauthors}%
Abbasi-Yadkori, Y.%
, P{\'a}l, D.%
\BCBL {}\ \BBA {} Szepesv{\'a}ri, C.%
\end{APACrefauthors}%
\unskip\
\newblock
\APACrefYearMonthDay{2011}{}{}.
\newblock
{\BBOQ}\APACrefatitle {Improved algorithms for linear stochastic bandits}
  {Improved algorithms for linear stochastic bandits}.{\BBCQ}
\newblock
\APACjournalVolNumPages{Advances in neural information processing
  systems}{24}{}{2312--2320}.
\PrintBackRefs{\CurrentBib}

\bibitem [\protect \citeauthoryear {%
Agrawal%
, Avadhanula%
, Goyal%
\BCBL {}\ \BBA {} Zeevi%
}{%
Agrawal%
\ \protect \BOthers {.}}{%
{\protect \APACyear {2016}}%
}]{%
agrawal2016near}
\APACinsertmetastar {%
agrawal2016near}%
\begin{APACrefauthors}%
Agrawal, S.%
, Avadhanula, V.%
, Goyal, V.%
\BCBL {}\ \BBA {} Zeevi, A.%
\end{APACrefauthors}%
\unskip\
\newblock
\APACrefYearMonthDay{2016}{}{}.
\newblock
{\BBOQ}\APACrefatitle {A near-optimal exploration-exploitation approach for
  assortment selection} {A near-optimal exploration-exploitation approach for
  assortment selection}.{\BBCQ}
\newblock
\BIn{} \APACrefbtitle {Proceedings of the 2016 ACM Conference on Economics and
  Computation} {Proceedings of the 2016 acm conference on economics and
  computation}\ (\BPGS\ 599--600).
\PrintBackRefs{\CurrentBib}

\bibitem [\protect \citeauthoryear {%
Agrawal%
, Avadhanula%
, Goyal%
\BCBL {}\ \BBA {} Zeevi%
}{%
Agrawal%
\ \protect \BOthers {.}}{%
{\protect \APACyear {2017}}%
}]{%
agrawal2017thompson}
\APACinsertmetastar {%
agrawal2017thompson}%
\begin{APACrefauthors}%
Agrawal, S.%
, Avadhanula, V.%
, Goyal, V.%
\BCBL {}\ \BBA {} Zeevi, A.%
\end{APACrefauthors}%
\unskip\
\newblock
\APACrefYearMonthDay{2017}{}{}.
\newblock
{\BBOQ}\APACrefatitle {Thompson sampling for the mnl-bandit} {Thompson sampling
  for the mnl-bandit}.{\BBCQ}
\newblock
\BIn{} \APACrefbtitle {Conference on Learning Theory} {Conference on learning
  theory}\ (\BPGS\ 76--78).
\PrintBackRefs{\CurrentBib}

\bibitem [\protect \citeauthoryear {%
Agrawal%
, Avadhanula%
, Goyal%
\BCBL {}\ \BBA {} Zeevi%
}{%
Agrawal%
\ \protect \BOthers {.}}{%
{\protect \APACyear {2019}}%
}]{%
agrawal2019mnl}
\APACinsertmetastar {%
agrawal2019mnl}%
\begin{APACrefauthors}%
Agrawal, S.%
, Avadhanula, V.%
, Goyal, V.%
\BCBL {}\ \BBA {} Zeevi, A.%
\end{APACrefauthors}%
\unskip\
\newblock
\APACrefYearMonthDay{2019}{}{}.
\newblock
{\BBOQ}\APACrefatitle {Mnl-bandit: A dynamic learning approach to assortment
  selection} {Mnl-bandit: A dynamic learning approach to assortment
  selection}.{\BBCQ}
\newblock
\APACjournalVolNumPages{Operations Research}{67}{5}{1453--1485}.
\PrintBackRefs{\CurrentBib}

\bibitem [\protect \citeauthoryear {%
Agrawal%
\ \BBA {} Devanur%
}{%
Agrawal%
\ \BBA {} Devanur%
}{%
{\protect \APACyear {2019}}%
}]{%
agrawal2019bandits}
\APACinsertmetastar {%
agrawal2019bandits}%
\begin{APACrefauthors}%
Agrawal, S.%
\BCBT {}\ \BBA {} Devanur, N\BPBI R.%
\end{APACrefauthors}%
\unskip\
\newblock
\APACrefYearMonthDay{2019}{}{}.
\newblock
{\BBOQ}\APACrefatitle {Bandits with global convex constraints and objective}
  {Bandits with global convex constraints and objective}.{\BBCQ}
\newblock
\APACjournalVolNumPages{Operations Research}{67}{5}{1486--1502}.
\PrintBackRefs{\CurrentBib}

\bibitem [\protect \citeauthoryear {%
Aznag%
, Goyal%
\BCBL {}\ \BBA {} Perivier%
}{%
Aznag%
\ \protect \BOthers {.}}{%
{\protect \APACyear {2021}}%
}]{%
aznag2021mnl}
\APACinsertmetastar {%
aznag2021mnl}%
\begin{APACrefauthors}%
Aznag, A.%
, Goyal, V.%
\BCBL {}\ \BBA {} Perivier, N.%
\end{APACrefauthors}%
\unskip\
\newblock
\APACrefYearMonthDay{2021}{}{}.
\newblock
{\BBOQ}\APACrefatitle {MNL-Bandit with Knapsacks} {Mnl-bandit with
  knapsacks}.{\BBCQ}
\newblock
\APACjournalVolNumPages{arXiv preprint arXiv:2106.01135}{}{}{}.
\PrintBackRefs{\CurrentBib}

\bibitem [\protect \citeauthoryear {%
Badanidiyuru%
, Kleinberg%
\BCBL {}\ \BBA {} Slivkins%
}{%
Badanidiyuru%
\ \protect \BOthers {.}}{%
{\protect \APACyear {2013}}%
}]{%
badanidiyuru2013bandits}
\APACinsertmetastar {%
badanidiyuru2013bandits}%
\begin{APACrefauthors}%
Badanidiyuru, A.%
, Kleinberg, R.%
\BCBL {}\ \BBA {} Slivkins, A.%
\end{APACrefauthors}%
\unskip\
\newblock
\APACrefYearMonthDay{2013}{}{}.
\newblock
{\BBOQ}\APACrefatitle {Bandits with knapsacks} {Bandits with knapsacks}.{\BBCQ}
\newblock
\BIn{} \APACrefbtitle {2013 IEEE 54th Annual Symposium on Foundations of
  Computer Science} {2013 ieee 54th annual symposium on foundations of computer
  science}\ (\BPGS\ 207--216).
\PrintBackRefs{\CurrentBib}

\bibitem [\protect \citeauthoryear {%
Cao%
, Rusmevichientong%
\BCBL {}\ \BBA {} Topaloglu%
}{%
Cao%
\ \protect \BOthers {.}}{%
{\protect \APACyear {2020}}%
}]{%
cao2020revenue}
\APACinsertmetastar {%
cao2020revenue}%
\begin{APACrefauthors}%
Cao, Y.%
, Rusmevichientong, P.%
\BCBL {}\ \BBA {} Topaloglu, H.%
\end{APACrefauthors}%
\unskip\
\newblock
\APACrefYearMonthDay{2020}{}{}.
\newblock
\APACrefbtitle {Revenue management under a mixture of multinomial logit and
  independent demand models} {Revenue management under a mixture of multinomial
  logit and independent demand models}\ \APACbVolEdTR{}{\BTR{}}.
\newblock
\APACaddressInstitution{}{Working paper}.
\PrintBackRefs{\CurrentBib}

\bibitem [\protect \citeauthoryear {%
Caro%
\ \BBA {} Gallien%
}{%
Caro%
\ \BBA {} Gallien%
}{%
{\protect \APACyear {2007}}%
}]{%
caro2007dynamic}
\APACinsertmetastar {%
caro2007dynamic}%
\begin{APACrefauthors}%
Caro, F.%
\BCBT {}\ \BBA {} Gallien, J.%
\end{APACrefauthors}%
\unskip\
\newblock
\APACrefYearMonthDay{2007}{}{}.
\newblock
{\BBOQ}\APACrefatitle {Dynamic assortment with demand learning for seasonal
  consumer goods} {Dynamic assortment with demand learning for seasonal
  consumer goods}.{\BBCQ}
\newblock
\APACjournalVolNumPages{Management science}{53}{2}{276--292}.
\PrintBackRefs{\CurrentBib}

\bibitem [\protect \citeauthoryear {%
Chen%
, Wang%
\BCBL {}\ \BBA {} Yuan%
}{%
Chen%
\ \protect \BOthers {.}}{%
{\protect \APACyear {2013}}%
}]{%
chen2013combinatorial}
\APACinsertmetastar {%
chen2013combinatorial}%
\begin{APACrefauthors}%
Chen, W.%
, Wang, Y.%
\BCBL {}\ \BBA {} Yuan, Y.%
\end{APACrefauthors}%
\unskip\
\newblock
\APACrefYearMonthDay{2013}{}{}.
\newblock
{\BBOQ}\APACrefatitle {Combinatorial multi-armed bandit: General framework and
  applications} {Combinatorial multi-armed bandit: General framework and
  applications}.{\BBCQ}
\newblock
\BIn{} \APACrefbtitle {International Conference on Machine Learning}
  {International conference on machine learning}\ (\BPGS\ 151--159).
\PrintBackRefs{\CurrentBib}

\bibitem [\protect \citeauthoryear {%
Cheung%
\ \BBA {} Simchi-Levi%
}{%
Cheung%
\ \BBA {} Simchi-Levi%
}{%
{\protect \APACyear {2017}}%
}]{%
cheung2017assortment}
\APACinsertmetastar {%
cheung2017assortment}%
\begin{APACrefauthors}%
Cheung, W\BPBI C.%
\BCBT {}\ \BBA {} Simchi-Levi, D.%
\end{APACrefauthors}%
\unskip\
\newblock
\APACrefYearMonthDay{2017}{}{}.
\newblock
{\BBOQ}\APACrefatitle {Assortment optimization under unknown multinomial logit
  choice models} {Assortment optimization under unknown multinomial logit
  choice models}.{\BBCQ}
\newblock
\APACjournalVolNumPages{arXiv preprint arXiv:1704.00108}{}{}{}.
\PrintBackRefs{\CurrentBib}

\bibitem [\protect \citeauthoryear {%
Cheung%
, Simchi-Levi%
\BCBL {}\ \BBA {} Wang%
}{%
Cheung%
\ \protect \BOthers {.}}{%
{\protect \APACyear {2017}}%
}]{%
cheung2017dynamic}
\APACinsertmetastar {%
cheung2017dynamic}%
\begin{APACrefauthors}%
Cheung, W\BPBI C.%
, Simchi-Levi, D.%
\BCBL {}\ \BBA {} Wang, H.%
\end{APACrefauthors}%
\unskip\
\newblock
\APACrefYearMonthDay{2017}{}{}.
\newblock
{\BBOQ}\APACrefatitle {Dynamic pricing and demand learning with limited price
  experimentation} {Dynamic pricing and demand learning with limited price
  experimentation}.{\BBCQ}
\newblock
\APACjournalVolNumPages{Operations Research}{65}{6}{1722--1731}.
\PrintBackRefs{\CurrentBib}

\bibitem [\protect \citeauthoryear {%
Davis%
, Gallego%
\BCBL {}\ \BBA {} Topaloglu%
}{%
Davis%
\ \protect \BOthers {.}}{%
{\protect \APACyear {2013}}%
}]{%
davis2013assortment}
\APACinsertmetastar {%
davis2013assortment}%
\begin{APACrefauthors}%
Davis, J.%
, Gallego, G.%
\BCBL {}\ \BBA {} Topaloglu, H.%
\end{APACrefauthors}%
\unskip\
\newblock
\APACrefYearMonthDay{2013}{}{}.
\newblock
{\BBOQ}\APACrefatitle {Assortment planning under the multinomial logit model
  with totally unimodular constraint structures} {Assortment planning under the
  multinomial logit model with totally unimodular constraint
  structures}.{\BBCQ}
\newblock
\APACjournalVolNumPages{Work in Progress}{}{}{}.
\PrintBackRefs{\CurrentBib}

\bibitem [\protect \citeauthoryear {%
D{\'e}sir%
, Goyal%
\BCBL {}\ \BBA {} Zhang%
}{%
D{\'e}sir%
\ \protect \BOthers {.}}{%
{\protect \APACyear {2014}}%
}]{%
desir2014near}
\APACinsertmetastar {%
desir2014near}%
\begin{APACrefauthors}%
D{\'e}sir, A.%
, Goyal, V.%
\BCBL {}\ \BBA {} Zhang, J.%
\end{APACrefauthors}%
\unskip\
\newblock
\APACrefYearMonthDay{2014}{}{}.
\newblock
{\BBOQ}\APACrefatitle {Near-optimal algorithms for capacity constrained
  assortment optimization} {Near-optimal algorithms for capacity constrained
  assortment optimization}.{\BBCQ}
\newblock
\APACjournalVolNumPages{Available at SSRN}{2543309}{}{}.
\PrintBackRefs{\CurrentBib}

\bibitem [\protect \citeauthoryear {%
Gallego%
, Iyengar%
, Phillips%
\BCBL {}\ \BBA {} Dubey%
}{%
Gallego%
\ \protect \BOthers {.}}{%
{\protect \APACyear {2004}}%
}]{%
gallego2004managing}
\APACinsertmetastar {%
gallego2004managing}%
\begin{APACrefauthors}%
Gallego, G.%
, Iyengar, G.%
, Phillips, R.%
\BCBL {}\ \BBA {} Dubey, A.%
\end{APACrefauthors}%
\unskip\
\newblock
\APACrefYearMonthDay{2004}{}{}.
\newblock
{\BBOQ}\APACrefatitle {Managing flexible products on a network} {Managing
  flexible products on a network}.{\BBCQ}
\newblock
\APACjournalVolNumPages{Available at SSRN 3567371}{}{}{}.
\PrintBackRefs{\CurrentBib}

\bibitem [\protect \citeauthoryear {%
Gao%
, Han%
, Ren%
\BCBL {}\ \BBA {} Zhou%
}{%
Gao%
\ \protect \BOthers {.}}{%
{\protect \APACyear {2019}}%
}]{%
gao2019batched}
\APACinsertmetastar {%
gao2019batched}%
\begin{APACrefauthors}%
Gao, Z.%
, Han, Y.%
, Ren, Z.%
\BCBL {}\ \BBA {} Zhou, Z.%
\end{APACrefauthors}%
\unskip\
\newblock
\APACrefYearMonthDay{2019}{}{}.
\newblock
{\BBOQ}\APACrefatitle {Batched multi-armed bandits problem} {Batched
  multi-armed bandits problem}.{\BBCQ}
\newblock
\APACjournalVolNumPages{arXiv preprint arXiv:1904.01763}{}{}{}.
\PrintBackRefs{\CurrentBib}

\bibitem [\protect \citeauthoryear {%
Kallus%
\ \BBA {} Udell%
}{%
Kallus%
\ \BBA {} Udell%
}{%
{\protect \APACyear {2020}}%
}]{%
kallus2020dynamic}
\APACinsertmetastar {%
kallus2020dynamic}%
\begin{APACrefauthors}%
Kallus, N.%
\BCBT {}\ \BBA {} Udell, M.%
\end{APACrefauthors}%
\unskip\
\newblock
\APACrefYearMonthDay{2020}{}{}.
\newblock
{\BBOQ}\APACrefatitle {Dynamic assortment personalization in high dimensions}
  {Dynamic assortment personalization in high dimensions}.{\BBCQ}
\newblock
\APACjournalVolNumPages{Operations Research}{68}{4}{1020--1037}.
\PrintBackRefs{\CurrentBib}

\bibitem [\protect \citeauthoryear {%
Kveton%
, Wen%
, Ashkan%
\BCBL {}\ \BBA {} Szepesvari%
}{%
Kveton%
\ \protect \BOthers {.}}{%
{\protect \APACyear {2015}}%
}]{%
kveton2015tight}
\APACinsertmetastar {%
kveton2015tight}%
\begin{APACrefauthors}%
Kveton, B.%
, Wen, Z.%
, Ashkan, A.%
\BCBL {}\ \BBA {} Szepesvari, C.%
\end{APACrefauthors}%
\unskip\
\newblock
\APACrefYearMonthDay{2015}{}{}.
\newblock
{\BBOQ}\APACrefatitle {Tight regret bounds for stochastic combinatorial
  semi-bandits} {Tight regret bounds for stochastic combinatorial
  semi-bandits}.{\BBCQ}
\newblock
\BIn{} \APACrefbtitle {Artificial Intelligence and Statistics} {Artificial
  intelligence and statistics}\ (\BPGS\ 535--543).
\PrintBackRefs{\CurrentBib}

\bibitem [\protect \citeauthoryear {%
Perchet%
, Rigollet%
, Chassang%
\BCBL {}\ \BBA {} Snowberg%
}{%
Perchet%
\ \protect \BOthers {.}}{%
{\protect \APACyear {2016}}%
}]{%
perchet2016batched}
\APACinsertmetastar {%
perchet2016batched}%
\begin{APACrefauthors}%
Perchet, V.%
, Rigollet, P.%
, Chassang, S.%
\BCBL {}\ \BBA {} Snowberg, E.%
\end{APACrefauthors}%
\unskip\
\newblock
\APACrefYearMonthDay{2016}{}{}.
\newblock
{\BBOQ}\APACrefatitle {Batched bandit problems} {Batched bandit
  problems}.{\BBCQ}
\newblock
\APACjournalVolNumPages{The Annals of Statistics}{44}{2}{660--681}.
\PrintBackRefs{\CurrentBib}

\bibitem [\protect \citeauthoryear {%
Rusmevichientong%
, Shen%
\BCBL {}\ \BBA {} Shmoys%
}{%
Rusmevichientong%
\ \protect \BOthers {.}}{%
{\protect \APACyear {2010}}%
}]{%
rusmevichientong2010dynamic}
\APACinsertmetastar {%
rusmevichientong2010dynamic}%
\begin{APACrefauthors}%
Rusmevichientong, P.%
, Shen, Z\BHBI J\BPBI M.%
\BCBL {}\ \BBA {} Shmoys, D\BPBI B.%
\end{APACrefauthors}%
\unskip\
\newblock
\APACrefYearMonthDay{2010}{}{}.
\newblock
{\BBOQ}\APACrefatitle {Dynamic assortment optimization with a multinomial logit
  choice model and capacity constraint} {Dynamic assortment optimization with a
  multinomial logit choice model and capacity constraint}.{\BBCQ}
\newblock
\APACjournalVolNumPages{Operations research}{58}{6}{1666--1680}.
\PrintBackRefs{\CurrentBib}

\bibitem [\protect \citeauthoryear {%
Saur{\'e}%
\ \BBA {} Zeevi%
}{%
Saur{\'e}%
\ \BBA {} Zeevi%
}{%
{\protect \APACyear {2013}}%
}]{%
saure2013optimal}
\APACinsertmetastar {%
saure2013optimal}%
\begin{APACrefauthors}%
Saur{\'e}, D.%
\BCBT {}\ \BBA {} Zeevi, A.%
\end{APACrefauthors}%
\unskip\
\newblock
\APACrefYearMonthDay{2013}{}{}.
\newblock
{\BBOQ}\APACrefatitle {Optimal dynamic assortment planning with demand
  learning} {Optimal dynamic assortment planning with demand learning}.{\BBCQ}
\newblock
\APACjournalVolNumPages{Manufacturing \& Service Operations
  Management}{15}{3}{387--404}.
\PrintBackRefs{\CurrentBib}

\bibitem [\protect \citeauthoryear {%
Simchi-Levi%
\ \BBA {} Xu%
}{%
Simchi-Levi%
\ \BBA {} Xu%
}{%
{\protect \APACyear {2019}}%
}]{%
simchi2019phase}
\APACinsertmetastar {%
simchi2019phase}%
\begin{APACrefauthors}%
Simchi-Levi, D.%
\BCBT {}\ \BBA {} Xu, Y.%
\end{APACrefauthors}%
\unskip\
\newblock
\APACrefYearMonthDay{2019}{}{}.
\newblock
{\BBOQ}\APACrefatitle {Phase transitions and cyclic phenomena in bandits with
  switching constraints} {Phase transitions and cyclic phenomena in bandits
  with switching constraints}.{\BBCQ}
\newblock
\APACjournalVolNumPages{Available at SSRN 3380783}{}{}{}.
\PrintBackRefs{\CurrentBib}

\bibitem [\protect \citeauthoryear {%
Talluri%
\ \BBA {} Van~Ryzin%
}{%
Talluri%
\ \BBA {} Van~Ryzin%
}{%
{\protect \APACyear {2004}}%
}]{%
talluri2004revenue}
\APACinsertmetastar {%
talluri2004revenue}%
\begin{APACrefauthors}%
Talluri, K.%
\BCBT {}\ \BBA {} Van~Ryzin, G.%
\end{APACrefauthors}%
\unskip\
\newblock
\APACrefYearMonthDay{2004}{}{}.
\newblock
{\BBOQ}\APACrefatitle {Revenue management under a general discrete choice model
  of consumer behavior} {Revenue management under a general discrete choice
  model of consumer behavior}.{\BBCQ}
\newblock
\APACjournalVolNumPages{Management Science}{50}{1}{15--33}.
\PrintBackRefs{\CurrentBib}

\end{thebibliography}

\appendix
\section{Some Important Facts}
	\begin{lemma}[Azuma-Hoeffding]\label{lemma:Azuma-Hoeffding}\cite{aznag2021mnl}
	 Consider random variables $X_{1}, \ldots, X_{n}$, defined with respect to a filtration $\mathcal{F}_{n}$ and a stopping time $\tau \leq n$ a.s, and that $\left(X_{k}\right)$ is uniformly bounded by $X^{*} .$ Then following inequality holds:
	$$
	\mathbb{P}\left(\sum_{k=\tau}^{n} X_{k}-\mathbb{E}\left(X_{k} \mid X_{k-1}, \ldots, X_{1}\right)>X^{*} \sqrt{2 n \log T}\right) \leq \frac{1}{T}
	$$
	\end{lemma}
	Note that the lemma is usually mentioned with a deterministic $\tau .$ However, notice that for each realization of $\tau$, one can derive a sharper bound and relax it using $0 \leq \tau \leq n .$ Since the probability does not depend on such a relization, a law of total probability gives the result.
	\begin{lemma}\label{lemma:7} \cite{abbasi2011improved}
	  Let $\left\{\mathcal{F}_{t}\right\}_{t=1}^{\infty}$ be a filtration. Let $\rho(t) \in\{0,1\}$ be a binary $\mathcal{F}_{t-1}$-measurable random variable, and let $\eta(t)$ be a $\mathcal{F}_{t}$-measurable random variable that is conditionally centered and $\mathcal{F}_{t-1}-$ conditionally $L$-subGaussian, i.e. $\mathbb{E}\left[\eta(t) \mid \mathcal{F}_{t-1}\right]=0$ a.s. and $\mathbb{E}\left[e^{\lambda \eta(t)} \mid \mathcal{F}_{t-1}\right] \leq e^{(\lambda L)^{2} / 2}$ for all $\lambda \in \mathbb{R}$. Then the confidence bound
	$$
	\left|\sum_{t=1}^{\tau} \rho(t) \eta(t)\right| \leq L \sqrt{\left(1+\sum_{t=1}^{\tau} \rho(t)\right)\left(1+2 \log \frac{\sqrt{(1+\sum_{t=1}^{\tau} \rho(t))}}{\delta}\right)}
	$$
	holds with probability at least $1-\delta$
	\end{lemma}
\section{Proofs}
\subsection{Proof of Lemma~\ref{lemma:event A}}
        The first-order derivative of $\mathcal{L}_{\ell-1}=-\sum_{t=1}^{T_{\ell-1}}\theta_{I_{t}}-\log(1+\sum_{i\in S_{t}}e^{\theta(i)})$ is:
	\begin{equation}\label{equation: first_derivative}
		\begin{aligned}
			\left.\frac{\partial \mathcal{L}_{\ell-1}}{\partial \theta(i)}\right|_{\theta=\theta^{*}} &=\sum_{t \in\{1, \ldots, T_{l-1}\}:i\in S_{t}} \varphi\left(i, S_{t} \mid v^{*}\right)-\mathbb{I}\left(I_{t}=i\right) \\
			&=\sum_{t=1}^{T_{l-1}} \rho_{t}(i) \left(\varphi\left(i, S_{t} \mid v^{*}\right)-\mathbb{I}\left(I_{t}=i\right)\right)
		\end{aligned}
	\end{equation}
where $\rho_{t}(i)=\mathbb{I}\left(S_{t} \ni i\right)$ is the indicator random variable of product $i$ being in the assortment $S_{t}$ in the $t^{\text {th }}$ period. Define the filtration $\mathcal{F}_{t-1}=\sigma\left(\left\{\left(S_{s}, I_{s}\right)\right\}_{s=1}^{t-1} \cup\right.$ $\left.\left\{S_{t}\right\}\right)$, the $\sigma$-algebra generated by $\left\{\left(S_{s}, I_{s}\right)\right\}_{s=1}^{t-1} \cup\left\{S_{t}\right\}$. Then the indicator $\rho_{t}(i)$ and the probability $\varphi\left(i, S_{t} \mid v^{*}\right)$ are $\mathcal{F}_{t-1}$-measurable, and the purchased product $I_{t}$ at period $t$ is $\mathcal{F}_{t}$-measurable. So, we have $\mathbb{E}\left[\mathbb{I}\left(I_{t}=i\right) \mid \mathcal{F}_{t-1}\right]=$ $\varphi\left(i, S_{t} \mid v^{*}\right)$. Clearly, $\varphi\left(i, S_{t} \mid v^{*}\right)-\mathbb{I}\left(I_{s}=i\right)$ is 1-subGaussian. Hence, based on the Lemma ~\ref{lemma:7}, we can get the bound of Eq.~\eqref{equation: first_derivative}\\
	\begin{equation}\label{equation:first_deriv_bound}
		\begin{aligned}\small
			\left|\frac{\partial \mathcal{L}_{\ell-1}}{\partial \theta(i)}\right|_{\theta=\theta^{*}} &=\left|\sum_{t=1}^{T_{\ell-1}} \rho_{t}(i)\left(\varphi\left(i, S_{t} \mid v^{*}\right)-\mathbb{I}\left(I_{t}=i\right)\right)\right| \\
			& \leq \sqrt{\left(1+n^{\ell-1}_i\right)\left(1+2 \log \frac{\sqrt{1+T_{\ell-1}}}{\delta_{1}}\right)} \\
			& \leq \sqrt{\left(1+n^{\ell-1}_i\right)\left(1+2 \log \frac{\sqrt{T}}{\delta_{1}}\right)} \\
			& \leq \sqrt{\left(2 n^{\ell-1}_i\right)\left(1+2 \log \frac{\sqrt{T}}{\delta_{1}}\right)} \\
		\end{aligned}
	\end{equation}
	holds with probability at least $1-\delta_{1} $ for each $i \in \mathcal{N}$, where the last inequality in the chain of Eq.~\eqref{equation:first_deriv_bound} holds, Because we show each product at least once during the warm-start period. We further define the following events:
    {\small $$
    	\mathcal{A}_{i,\ell}=\left\{\left|\frac{\partial L_{\ell-1}}{\partial \theta(i)}\right|_{\theta=\theta^{*}}  \leq \sqrt{\left(2 n^{\ell-1}_i\right)\left(1+2 \log \frac{2\sqrt{T}q(L,K)N}{\delta}\right)}\right\}\\
    $$
    $$
    	\mathcal{A}_{\ell}=\bigcap_{i=1}^{N}\mathcal{A}_{i,\ell},
    	\mathcal{A}=\bigcap_{i=1}^{N}\bigcap_{\ell=1}^{q(L,K)}\mathcal{A}_{i,\ell}
    $$}
    Therefore, when setting $\delta_{1}=1-\delta/2q(L,K)N$, we get $\mathbb{P}_{\pi}\left(\mathcal{A}_{i, \ell}\right) \geq 1-\delta/2q(L,K)N$, which bound the first derivative of $\theta(i)$ at each epoch $\ell\in [q(L,K)]$. Combing with union bound, we get $\mathbb{P}_{\pi}\left(\mathcal{A}_{ \ell}\right) \geq 1-\delta/2q(L,K)$, and $\mathbb{P}_{\pi}\left(\mathcal{A}\right) \geq 1-\delta/2$.\\

\subsection{ Proof of Theorem ~\ref{theorem:parameter}}

In this part, we bound the estimated parameters and true parameters. By Taylor approximation, we know that there exists $\gamma \in$ $[0,1]$ such that
	\begin{equation}\label{equation:taylor_approx}\small
			\mathcal{L}_{\ell-1}\left(\theta_{l}\right)=\mathcal{L}_{\ell-1}\left(\theta^{*}\right)+\nabla \mathcal{L}_{\ell-1}\left(\theta^{*}\right)^{T}\left(\theta_{\ell}-\theta^{*}\right) +\frac{1}{2}\left(\theta_{\ell}-\theta^{*}\right)^{T} H_{\ell-1}\left(\theta^{*}+\gamma\left(\theta_{\ell}-\theta^{*}\right)\right)\left(\theta_{\ell}-\theta^{*}\right)
	\end{equation}
	where the first gradient and hessian matrix are:
	\begin{equation}\label{equation:firstorder}\small
			\nabla 	\mathcal{L}_{\ell-1}(\theta^{*})=\left.\left(\frac{\partial \mathcal{L}_{\ell-1}(\theta)}{\partial \theta(i)}\right)_{i=1}^{N}\right|_{\theta=\theta^{*}}
	\end{equation}
	\begin{equation}\label{equation:hess}\small
			H_{\ell-1}\left(\theta^{*}+\gamma\left(\theta_{\ell}-\theta^{*}\right)\right)=\left.\left(\frac{\partial^{2} \mathcal{L}_{\ell-1}(\theta)}{\partial \theta(i) \partial \theta(j)}\right)_{1 \leq i, j \leq N}\right|_{\theta=\theta^{*}+\gamma\left(\theta_{\ell}-\theta^{*}\right)}
	\end{equation}
	Since $\hat{\theta}_{\ell}$ minimizes the $\mathcal{L}_{\ell-1}$, thus we have
	\begin{equation}\label{equation:hat_theta}\small
			\mathcal{L}_{\ell-1}(\theta^{*})\ge \mathcal{L}_{\ell-1}(\hat{\theta}_{\ell})
	\end{equation}
	Further,
	\begin{equation}\small
			\nabla \mathcal{L}_{\ell-1}\left(\theta^{*}\right)^{T}\left(\hat{\theta}_{\ell}-\theta^{*}\right) +\frac{1}{2}\left(\hat{\theta}_{\ell}-\theta^{*}\right)^{T} H_{\ell-1}\left(\theta^{*}+\gamma\left(\hat{\theta}_{\ell}-\theta^{*}\right)\right)\left(\hat{\theta}_{\ell}-\theta^{*}\right)\le 0
	\end{equation}
	To bound the estimated parameters, the following two derivatives are given.
	\begin{equation}\small
			\frac{\partial^{2} \mathcal{L}_{\ell-1}}{\partial \theta(i) \partial \theta(j)}=-\sum_{t \in\{1, \cdots, T_{\ell-1}\}:S_{t}\ni i,j}\frac{e^{\theta(i)}e^{\theta(j)}}{\left(1+\sum_{k \in S_{t}} e^{\theta(k)}\right)^{2}} \cdot
	\end{equation}
	\begin{equation}\small
		\begin{aligned}					
			\frac{\partial^{2} \mathcal{L}_{\ell-1}}{\partial \theta(i)^{2}}=-\sum_{t \in\{1, \cdots, T_{\ell-1}\}:S_{t}\ni i}\frac{e^{\theta(i)}+\sum_{k \in S_{t} \backslash i} e^{\theta(i)} e^{\theta(k)}}{\left(1+\sum_{k \in S_{t}} e^{\theta(k)}\right)^{2}} \cdot
		\end{aligned}
	\end{equation}
	Hence, the Hessian matrix $h_{t}(\theta)$ for the $t^{th}$ sample is as follow:
	\begin{equation}
		\begin{aligned}\small
			&h_{t}(\theta)=\frac{1}{\left(1+\sum_{i \in S_{t}} e^{\theta(i)}\right)^{2}} \times \\
			&\left(\begin{array}{ccccc}
				e^{\theta(1)} 1\left(1 \in S_{t}\right) & 0 & 0 & 0 & 0 \\
				0 & e^{\theta(2)} 1\left(2 \in S_{t}\right) & 0 & 0 & 0 \\
				0 & 0 & \ddots & 0 & 0 \\
				0 & 0 & 0 & \ddots & 0 \\
				0 & 0 & 0 & 0 & e^{\theta(N)} 1\left(N \in S_{t}\right)
			\end{array}\right) \\
			& +\sum_{1 \leq i<j \leq N: \atop i, j \in S_{t}} e^{\theta(i)+\theta(j)} u_{i, j} u_{i, j}^{T},
		\end{aligned}
	\end{equation}
	where the vector $u_{i, j}=e_{i}-e_{j}$, and $e_{i}$ is the $i^{\text {th }}$ standard basis vector. Now, each term in the second summation is positive semi-definite. Applying the bound $v=e^{\theta(i)} \in[1/R,R]$ for all $i \in \mathcal{N}$, and in this model $|S| \leq N$ for all $S \in \mathcal{S}$, we have
	\begin{equation}\label{hess:h(theta)}
		\begin{aligned}
			&h_{t}(\theta) \succeq \frac{1}{R(1+N R)^{2}} \times \\
			&\left(\begin{array}{ccccc}
				1\left(1 \in S_{t}\right) & 0 & 0 & 0 & 0 \\
				0 & 1\left(2 \in S_{t}\right) & 0 & 0 & 0 \\
				0 & 0 & \ddots & 0 & 0 \\
				0 & 0 & 0 & \ddots & 0 \\
				0 & 0 & 0 & 0 & 1\left(N \in S_{t}\right)
			\end{array}\right)
		\end{aligned}
	\end{equation}
	Summing the Eq.~\eqref{hess:h(theta)} over $1 \leq t \leq T_{\ell-1}$ yields Eq. ~\eqref{hess:H(theta)}. 
	\begin{equation}\label{hess:H(theta)}
		\begin{aligned}
			&H_{\ell-1}(\theta) \succeq \frac{1}{R(1+NR)^{2}} \times \\
			&\left(\begin{array}{ccccc}
				n^{\ell-1}_1 & 0 & 0 & 0 & 0 \\
				0 &n^{\ell-1}_2 & 0 & 0 & 0 \\
				0 & 0 & \ddots & 0 & 0 \\
				0 & 0 & 0 & \ddots & 0 \\
				0 & 0 & 0 & 0 & n^{\ell-1}_N
			\end{array}\right)
		\end{aligned}
	\end{equation}
	Therefore, under event $\mathcal{A}$, we have
	{\small $$
	\frac{1}{R(1+N R)^{2}} \sum_{i=1}^{N}\left(\sqrt{n^{\ell-1}_i}\left(\hat{\theta}_{\ell}(i)-\theta^{*}(i)\right)\right)^{2}
	-\sqrt{\left(2+4 \log \frac{2T^{1/2}q(L,K)N}{\delta}\right)} \sum_{i=1}^{N} \sqrt{n^{\ell-1}_i}\left|\hat{\theta}_{\ell}(i)-\theta^{*}(i)\right| \leq 0
	$$
	$$
	\sum_{i=1}^{N}\left(\sqrt{n^{\ell-1}_i}\left(\hat{\theta}_{\ell}(i)-\theta^{*}(i)\right)\right)^{2}-2 \Psi \sum_{i=1}^{N} \sqrt{n^{\ell-1}_i}\left|\hat{\theta}_{\ell}(i)-\theta^{*}(i)\right|+\sum_{i=1}^{N} \Psi^{2} \leq N \Psi^{2}
	$$
	$$
	\sum_{i=1}^{N}\left(\sqrt{n^{\ell-1}_i}\left(\hat{\theta}_{\ell}(i)-\theta^{*}(i)\right)-\Psi\right)^{2} \leq N \Psi^{2}
	$$}
	Finally, we derive the bound as:
	$$
	\left|\log \frac{\hat{v}_{i}^{\ell}}{v_{i}^{*}}\right| \leq \frac{(\sqrt{N}+1) \Psi}{\sqrt{n^{\ell-1}_i}}=	\varepsilon(n^{\ell-1}_i)
	$$
	where $\Psi=\frac{R(1+N R)^{2}}{2}\sqrt{\left(2+4 \log \frac{2T^{1/2}q(L,K)N}{\delta}\right)}$.

    \subsection{ Proof of Lemma ~\ref{lemma:lipschitz}}

        	Consider function $f:[0,1] \rightarrow \mathbb{R}$ defined by $f(\gamma)=\varphi\left(i, S \mid \exp \left[\theta^{\prime}+\gamma\left(\theta-\theta^{\prime}\right)\right]\right)$, where $\theta^{\prime}=$ $\left(\theta^{\prime}(i)\right)_{i \in \mathcal{N}}=\left(\log \left[v^{\prime}(i)\right]\right)_{i \in \mathcal{N}}$, and $\theta=(\theta(i))_{i \in \mathcal{N}}=$ $(\log [v(i)])_{i \in \mathcal{N}}$. Let's also define the shorthand $\theta_{\gamma}(i)=\theta^{\prime}(i)+\gamma\left(\theta(i)-\theta^{\prime}(i)\right)$. Note that $\theta_{0}=\theta^{\prime}$ and $\theta_{1}=\theta$. By the mean value theorem,
    \begin{align*}
    & \varphi(i, S \mid v)-\varphi\left(i, S \mid v^{\prime}\right) \\
    =& f(1)-f(0)=f^{\prime}(\gamma) \quad \text { for some } \gamma \in(0,1) \\
    =& \frac{e^{\theta_{\gamma}(i)}}{1+\sum_{\ell \in S} e^{\theta_{\gamma}(\ell)}}\left(\theta(i)-\theta^{\prime}(i)\right) -\frac{e^{2 \theta_{\gamma}(i)} }{\left(1+\sum_{\ell \in S} e^{\theta_{\gamma}(\ell)}\right)^{2}}\left(\theta(i)-\theta^{\prime}(i)\right) \\
    \leq &\left|\theta(i)-\theta^{\prime}(i)\right|\\
    =&\left|\log \frac{v(i)}{v^{\prime}(i)}\right|,
    \end{align*}
    where the last inequality holds since the sum of the coefficients of $\theta(i)-\theta^{\prime}(i)$ in the two summations lies in [0, 1]. Hence, we have
    $$
    \sum_{i \in S} b(i)\left(\varphi(i, S \mid v)-\varphi\left(i, S \mid v^{\prime}\right)\right) 
    \leq  \sum_{i \in S}b(i)\left|\theta(i)-\theta^{\prime}(i)\right|=\sum_{i \in S}b(i)\left|\log \frac{v(i)}{v^{\prime}(i)}\right|
    $$
    Together with Lemma~\ref{lemma:lipschitz} and Theorem~\ref{theorem:parameter}, the Corollary~\ref{corollary:1} can also be proved.

\subsection{ Proof of Lemma ~\ref{lemma:ResourceBound}}
For each resource $k \in \mathcal{K}$, the total consumed quantity over the $T$ periods can be calculated as follows:
	\begin{subequations}\small
		\begin{align}
			&\quad\sum_{t=T_{0}+1}^{T} a\left(I_{t},k\right)\notag\\	&=\sum_{\ell=1}^{q(L,K)}\sum_{t=T_{\ell-1}+1}^{T_{\ell}} a\left(I_{t},k\right)-\sum_{\ell=1}^{q(L,K)}\sum_{t=T_{\ell-1}+1}^{T_{\ell}} \left(\sum_{i \in S_{t}} a(i,k) \varphi(i, S_{t} \mid v^{*})\right)\label{re_constraint:1}\\
			&+\sum_{\ell=1}^{q(L,K)}\sum_{t=T_{\ell-1}+1}^{T_{\ell}}\left(\sum_{i \in S_{t}} a(i,k) \varphi(i, S_{t} \mid v^{*})\right) -\sum_{\ell=1}^{q(L,K)}\sum_{t=T_{\ell-1}+1}^{T_{\ell}} \left(\sum_{i \in S_{t}} a(i,k) \Big(\varphi(i, S_{t} \mid \hat{v}^{\ell})-\epsilon(n^{\ell-1}_i)\Big)\right)\label{re_constraint:2}\\
			&+\sum_{\ell=1}^{q(L,K)}\sum_{t=T_{\ell-1}+1}^{T_{\ell}} \left(\sum_{i \in S_{t}} a(i,k) \Big(\varphi(i, S_{t} \mid \hat{v}^{\ell})-\epsilon(n^{\ell-1}_i)\Big)-\sum_{S\in \mathcal{S}}y_{\ell}(S)\sum_{i \in S} a(i,k) \Big(\varphi(i, S \mid \hat{v}^{\ell})-\epsilon(n^{\ell-1}_i)\Big)\right)\label{re_constraint:3}\\
			&+\sum_{\ell=1}^{q(L,K)}\sum_{t=T_{\ell-1}+1}^{T_{\ell}}\sum_{S\in \mathcal{S}} y_{\ell}(S) \left( \sum_{i \in S} a(i,k) \Big(\varphi(i, S \mid \hat{v}^{\ell})-\epsilon(n^{\ell-1}_i)\Big) \right) \label{re_constraint:4}
		\end{align}
	\end{subequations}
	
    \textbf{First, we bound (\ref{re_constraint:2}).}
	\begin{subequations}\small
		\begin{align}
			&\quad\sum_{\ell=1}^{q(L,K)}\sum_{t=T_{\ell-1}+1}^{T_{\ell}} \left(\sum_{i \in S_{t}} a(i,k) \varphi(i, S_{t} \mid v^{*})\right)-\sum_{\ell=1}^{q(L,K)}\sum_{t=T_{\ell-1}+1}^{T_{\ell}} \left(\sum_{i \in S_{t}} a(i,k) (\varphi(i, S_{t} \mid \hat{v}^{\ell})-\epsilon(n^{\ell-1}_i))\right)\notag\\
			&\leq2\sum_{\ell=1}^{q(L,K)}\sum_{i=1}^{N}\sum_{j=1}^{N_{\ell}(i)}\epsilon(n^{\ell-1}_i)\notag\\
			&\leq2(\sqrt{N}+1) \Psi	\sum_{l=1}^{q(L,K)}\sum_{i=1}^{N}\sum_{j=1}^{N_{\ell}(i)}\frac{1}{\sqrt{n^{\ell-1}_i+j-1}} \underbrace{\sum_{j=1}^{N_{\ell}(i)}\frac{1}{\sqrt{n^{\ell-1}_i}}/\sum_{j=1}^{N_{\ell}(i)}\frac{1}{\sqrt{n^{\ell-1}_i+j-1}}}_{\left(\diamond_{0}\right)}\label{equation:frac}
		\end{align}
	\end{subequations}
	We derive the last inequality of Eq.~\eqref{equation:frac} based on the definition of $\epsilon(n^{\ell-1}_i)$, and decompose it into a special structure. Now, we focus on the part $\diamond_{0}$.
	\begin{equation}\label{equation:driv_frac}
		\begin{split}
			\diamond_{0}&=\sum_{j=1}^{N_{\ell}(i)}\frac{1}{\sqrt{n^{\ell-1}_i}}/\sum_{j=1}^{N_{\ell}(i)}\frac{1}{\sqrt{n^{\ell-1}_i+j-1}}\\
			&=N_{\ell}(i)/\sum_{j=1}^{N_{\ell}(i)}\frac{\sqrt{n^{\ell-1}_i}}{\sqrt{n^{\ell-1}_i+j-1}}\\
			&\leq \frac{\sqrt{n^{\ell-1}_i+N_{\ell}(i)-1}}{\sqrt{n^{\ell-1}_i}} \leq \sqrt{1+\frac{N_{\ell}(i)}{n^{\ell-1}_i}}
		\end{split}
	\end{equation}
	When $N_{\ell}(i)=T/q(L,K)$ and $n^{\ell-1}_i=\tau/N$, the $\diamond_{0}$ maximum. Hence,
	\begin{equation}\label{eq38}
			\quad\sum_{j=1}^{N_{\ell}(i)}\frac{1}{\sqrt{n^{\ell-1}_i}}/\sum_{j=1}^{N_{\ell}(i)}\frac{1}{\sqrt{n^{\ell-1}_i+j-1}}\leq\sqrt{1+\frac{NT}{\tau q(L,K)}}
	\end{equation}
    Thus, the bound of (\ref{re_constraint:2}) can be represented as:
	\begin{subequations}\small
		\begin{align}
			&\quad\sum_{\ell=1}^{q(L,K)}\sum_{t=T_{\ell-1}+1}^{T_{\ell}} \left(\sum_{i \in S_{t}} a(i,k) \varphi(i, S_{t} \mid v^{*})\right)-\sum_{\ell=1}^{q(L,K)}\sum_{t=T_{\ell-1}+1}^{T_{\ell}} \left(\sum_{i \in S_{t}} a(i,k) (\varphi(i, S_{t} \mid \hat{v}_{\ell})-\epsilon(n^{\ell-1}_i))\right)\notag\\
			&\leq2(\sqrt{N}+1)\sqrt{1+\frac{NT}{\tau q(L,K)}}
			\Psi\sum_{i=1}^{N}\sum_{\ell=1}^{q(L,K)}\sum_{j=1}^{N_{\ell}(i)}\frac{1}{\sqrt{n^{\ell-1}_i+j-1}}\notag\\
			&\leq4(\sqrt{N}+1)\sqrt{1+\frac{NT}{\tau q(L,K)}}\Psi\sum_{i=1}^{N}\sqrt{n_{T}(i)}\label{equation:inequality}\\
			&\leq4(\sqrt{N}+1)\sqrt{1+\frac{NT}{\tau q(L,K)}}\Psi\sqrt{N\sum_{i=1}^{N}n_{T}(i)}\label{equation:JensonInequality}\\
			&\leq4(\sqrt{N}+1)\sqrt{1+\frac{NT}{\tau q(L,K)}}\Psi\sqrt{N^{2}T}\label{equation:capacity}
		\end{align}
	\end{subequations}
Eq.~\eqref{equation:inequality} is based on the inequality:$\sum_{j=1}^{N}\frac{1}{\sqrt{j}}\leq 2\sqrt{N}$ and Eq.~\eqref{equation:JensonInequality} is by Jensen’s Inequality. Finally, the Eq.~\eqref{equation:capacity} is by the fact that at most $N$ products can be included in each of the $T$ periods.\\

\textbf{Then, we bound (\ref{re_constraint:1}) and (\ref{re_constraint:3}).}
According to Eq.~\eqref{equation:epsilon}, we have $$\left|a\left(I_{t},k\right)-\sum_{i \in S_{t}} a(i,k) \varphi(i, S_{t} \mid v^{*})\right|\leq 1$$ 
and 
$$\left|\sum_{i \in S} a(i, k) \big[\varphi\left(i, S \mid \hat{v}^{\ell}\right)-\varepsilon\left(n_{T_{\ell-1}}(i)\right)\big]\right| \leq \frac{2N^{2} \Psi}{\sqrt{\tau}}$$ 
for all $i \in S \in \mathcal{S}$ and all $\ell$. We also observe that 
$$U_{t}^{1}=a\left(I_{t},k\right)-\sum_{i \in S_{t}} a(i,k) \varphi(i, S_{t} \mid v^{*})$$ 
and 
$$U_{t}^{2}=\sum_{i \in S_{t}} a(i, k) \Big[\varphi\left(i, S_{t} \mid \hat{v}^{\ell}\right)-\varepsilon\left(n_{T_{\ell-1}}(i)\right)\Big]-\sum_{S \in \mathcal{S}}\Big[\sum_{i \in S} a(i, k) (\varphi\left(i, S \mid \hat{v}^{\ell}\right)-\varepsilon\left(n_{T_{l-1}}(i))\right)\Big] y_{l}(S)$$ both are martingale differences with respect to the filtration $\mathcal{E}_{t}=\sigma\left(\left\{S_{s}, I_{s}\right)\right\}_{s=1}^{t}$. By applying Azuma-Hoeffding inequality shown in Lemma~\ref{lemma:Azuma-Hoeffding}, we have
\begin{equation}\label{equation:P_1}\small
	 \mathbb{P}_{\pi}\left(\sum_{\ell=1}^{q(L,K)}\sum_{t=T_{\ell-1}+1}^{T_{\ell}} U_{t}^{1}>\sqrt{2T\log\frac{4(K+1)}{\delta}}\right)\leq \frac{\delta}{4(K+1)}
\end{equation}	

\begin{equation}\label{equation:P_2}\small
	 \mathbb{P}_{\pi}\left(\sum_{\ell=1}^{q(L,K)}\sum_{t=T_{\ell-1}+1}^{T_{\ell}} U_{t}^{2}>\frac{2N^{2}\Psi}{\sqrt{\tau}}\sqrt{2T\log\frac{4(K+1)}{\delta}}\right)\leq \frac{\delta}{4(K+1)}
\end{equation}
	
\textbf{Based on the constraints of Eq. \eqref{model:Choice_LP}, we have}
\begin{equation}\label{equation:P_3}
		\begin{aligned}
			\sum_{\ell=1}^{q(L,K)}\sum_{t=T_{\ell-1}+1}^{T_{\ell}}\sum_{S\in\mathcal{S}}y_{\ell}(S)\left(\sum_{i \in S} a(i,k) (\varphi(i, S \mid \hat{v}^{\ell})-\epsilon(n^{\ell-1}_i))\right)\leq T(1-w)c(k)
		\end{aligned}
\end{equation}

\textbf{The total amount $\sum_{t=1}^{T} a\left(I_{t}, k\right)$ of resource $k$ consumed from period 1 to period $T$ is at most}
	\begin{equation}
		\begin{split}
			& \tau+\eqref{equation:capacity}+\eqref{equation:P_1}+\eqref{equation:P_2}+\eqref{equation:P_3}\\
			<&\tau+4(\sqrt{N}+1)\sqrt{1+\frac{NT}{\tau q(L,K)}}\Psi\sqrt{N^{2}T}+\sqrt{2T\log\frac{4(K+1)}{\delta}}+\frac{2N^{2} \Psi}{\sqrt{\tau}} \sqrt{2T\log\frac{4(K+1)}{\delta}}\\
			&+T(1-w)c(k) \\
			<& Tc(k)
		\end{split}
	\end{equation}
    Therefore, if we set
	{\small $$
	\omega=\frac{1}{T\min _{k \in \mathcal{K}} c(k)}\left( \tau+4(\sqrt{N}+1)\sqrt{1+\frac{NT}{\tau q(L,K)}}\Psi\sqrt{N^{2}T}+\sqrt{2T\log\frac{4(K+1)}{\delta}}+\frac{2N^{2} \Psi}{\sqrt{\tau}} \sqrt{2T\log\frac{4(K+1)}{\delta}}\right),
    $$}
	the resource $k$ is not depleted with probability at least $1-\delta /2(K+1)$. Thus, all resources cannot be depleted with probability at least $1-K \delta /2(K+1)$.

\subsection{Proof of Lemma ~\ref{lemma:4}}
Let $y^{*}$ be an optimal solution to $\operatorname{LP}\left(v^{*}\right)$, and consider the solution $\bar{y}=(1-\omega) y^{*}+\omega 1_{\emptyset}$. That is $\bar{y}(S)=(1-\omega) y^{*}(S)$ for $S \in \mathcal{S} \backslash\{\emptyset\}$, and $\bar{y}(\emptyset)=(1-\omega) y^{*}(S)+\omega$. First, it is clear that $\bar{y}$ is feasible to UCB-LP $\left(\hat{v}^{\ell}, \hat{n}_{\ell-1}, \omega\right)$. Clearly, $\bar{y} \geq 0$, and $\sum_{S \in \mathcal{S}} \bar{y}(S)=(1-\omega) \sum_{S \in \mathcal{S}} y^{*}(S)+\omega=1 .$ Moreover, for each resource $k \in \mathcal{K}$, we have:
	\begin{equation}\small
		\begin{aligned}
			&\quad\sum_{S \in \mathcal{S}}\left(\sum_{i \in S} a(i, k)\big[\varphi\left(i, S \mid \hat{v}^{\ell}\right)-\varepsilon(n^{\ell-1}_i)\big]\right) \bar{y}(S)\\ 
			&=(1-\omega) \sum_{S \in \mathcal{S}}\left(\sum_{i \in S} a(i, k)\big[ \varphi\left(i, S \mid \hat{v}^{\ell}\right)-\varepsilon(n^{\ell-1}_i)\big]\right) y^{*}(S) \\
			&\leq(1-\omega) \sum_{S \in \mathcal{S}}\left(\sum_{i \in S} a(i, k) \varphi\left(i, S \mid v^{*}\right)\right) y^{*}(S) \leq(1-\omega) c(k)
		\end{aligned}
	\end{equation}
	For the objective function, suppose $\hat{y}_{\ell}(S)$ is optimal for UCB-LP$(\hat{v}_{\ell},n^{\ell-1},\omega)$we have
	\begin{equation}\small
    \begin{aligned}
        & \operatorname{OPT}\left(\operatorname{UCB}-\operatorname{LP}\left(\hat{v}^{\ell}, n^{\ell-1}, \omega\right)\right) \\
        =& \sum_{S \in \mathcal{S}}\left(\sum_{i \in S} r(i) \big[\varphi\left(i, S \mid \hat{v}^{\ell}\right)+\varepsilon(n^{\ell-1}_i)\big]\right) \hat{y}_{\ell}(S) \\
        \geq & \sum_{S \in \mathcal{S}}\left(\sum_{i \in S} r(i)\big[ \varphi\left(i, S \mid \hat{v}^{\ell}\right)+\varepsilon(n^{\ell-1}_i)\big]\right) \bar{y}(S) \geq \sum_{S \in \mathcal{S}} \sum_{i \in S} r(i) \varphi\left(i, S \mid v^{*}\right) \bar{y}(S) =(1-\omega) \operatorname{OPT}\left(\operatorname{LP}\left(v^{*}\right)\right).
    \end{aligned}
    \end{equation}

\subsection{ Proof of Lemma~\ref{lemma:Regretbound}}
The proof of Lemma~\ref{lemma:Regretbound} is similar to the proof of Lemma~\ref{lemma:ResourceBound}. We give a brief description. We decompose the regret into several parts
	\begin{subequations}\small
		\begin{align}
			&\quad (1-\omega)(T-\tau) \text{OPT}(\text{LP}(v^{*}))-\sum_{t=T_{0}+1}^{T} r\left(I_{t}\right) \notag\\
			&\leq\sum_{\ell=1}^{q(L,K)}\sum_{t=T_{\ell-1}+1}^{T_{\ell}}\sum_{S\in \mathcal{S}}y_{\ell}(S)\left(\sum_{i \in S} r(i) \big[\varphi(i, S \mid \hat{v}^{\ell})+\epsilon(n^{\ell-1}_i)\big]\right)-\sum_{t=\tau+1}^{T} r\left(I_{t}\right)\label{reg_constraint:1} \\
			&\leq\sum_{\ell=1}^{q(L,K)}\sum_{t=T_{\ell-1}+1}^{T_{\ell}}\sum_{S\in \mathcal{S}}y_{\ell}(S)\left(\sum_{i \in S} r(i) \big[\varphi(i, S \mid \hat{v}^{\ell})+\epsilon(n^{\ell-1}_i)\big]\right)-\sum_{\ell=1}^{q(L,K)}\sum_{t=T_{\ell-1}+1}^{T_{\ell}}\left(\sum_{i \in S_{t}} r(i)\big[\varphi(i, S_{t} \mid \hat{v}^{\ell})+\epsilon(n^{\ell-1}_i)\big]\right)\label{reg_constraint:2}\\
			&+\sum_{\ell=1}^{q(L,K)}\sum_{t=T_{\ell-1}+1}^{T_{\ell}}\left(\sum_{i \in S_{t}} r(i)\big[\varphi(i, S_{t} \mid \hat{v}^{\ell})+\epsilon(n^{\ell-1}_i)\big]\right)-\sum_{\ell=1}^{q(L,K)}\sum_{t=T_{\ell-1}+1}^{T_{\ell}}\left(\sum_{i \in S_{t}} r(i)\varphi(i, S_{t} \mid v^{*})\right)\label{reg_constraint:3}\\
			&+\sum_{\ell=1}^{q(L,K)}\sum_{t=T_{\ell-1}+1}^{T_{\ell}}\left(\sum_{i \in S_{t}} r(i)\varphi(i, S_{t} \mid v^{*})\right)-\sum_{t=\tau+1}^{T} r\left(I_{t}\right)\label{reg_constraint:4}
		\end{align}
	\end{subequations}

To bound (~\ref{reg_constraint:4}),
\begin{equation*}
	\mathbb{P}_{\pi}\left(\left(   \sum_{\ell=1}^{q(L,K)}\sum_{t=T_{\ell-1}+1}^{T_{\ell}}\left(\sum_{i \in S_{t}} r(i)\varphi(i, S_{t} \mid v^{*})\right)-\sum_{t=\tau+1}^{T} r\left(I_{t}\right)
	\right)>\sqrt{2T\log\frac{4(K+1)}{\delta}}\right)\leq \frac{\delta}{4(K+1)}
\end{equation*}	

We denote $\sum_{S\in \mathcal{S}}y_{\ell}(S)\left(\sum_{i \in S} r(i) \left(\varphi(i, S \mid \hat{v}^{\ell})+\epsilon(n^{\ell-1}_i)\right)\right)-\sum_{i \in S_{t}} r(i)\left(\varphi(i, S_{t} \mid \hat{v}^{\ell})+\epsilon(n^{\ell-1}_i)\right)$ as $U_{t}^{3}$, then the bound of Eq.~\eqref{reg_constraint:2} can be presented as:
\begin{equation}
	    	\quad\mathbb{P}_{\pi}\left(
	    	\sum_{\ell=1}^{q(L,K)}\sum_{t=T_{\ell-1}+1}^{T_{\ell}} U_{t}^{3}>\left(\frac{2N^{2} \Psi}{\sqrt{\tau}}+N\right) \sqrt{2T\log\frac{4(K+1)}{\delta}}\right) \leq \frac{\delta}{4(K+1)}
\end{equation}

To Bound (\ref{reg_constraint:3})
\begin{equation}
		\begin{split}		
			&\sum_{\ell=1}^{q(L,K)}\sum_{t=T_{\ell-1}+1}^{T_{\ell}}\left(\sum_{i \in S_{t}} r(i)\big[\varphi(i, S_{t} \mid \hat{v}^{\ell})+\epsilon(n^{\ell-1}_i)\big]\right)-\sum_{\ell=1}^{q(L,K)}\sum_{t=T_{\ell-1}+1}^{T_{\ell}}\left(\sum_{i \in S_{t}} r(i)\varphi(i, S_{t} \mid v_{*})\right)\notag\\
			&\leq4(\sqrt{N}+1)\sqrt{1+\frac{NT}{\tau q(L,K)}}\Psi\sqrt{N^{2}T}
		\end{split}
\end{equation}	
	
Put the above three inequalities together, we have
	\begin{equation}
		\begin{split}	
			&\quad(1-\omega)(T-T_{0}) \text{OPT}(\text{LP}(v^{*}))-\sum_{t=T_{0}+1}^{T} r\left(I_{t}\right)\\
			&\leq 4(\sqrt{N}+1)\sqrt{1+\frac{NT}{\tau q(L,K)}}\Psi\sqrt{N^{2}T}+\sqrt{2T\log\frac{4(K+1)}{\delta}}+\left(\frac{2N^{2} \Psi}{\sqrt{\tau}}+N \right)\sqrt{2T\log\frac{4(K+1)}{\delta}}
		\end{split}
	\end{equation}
	We have the regret as
	\begin{equation}\small
	    \begin{split}
        	&\quad(1-\omega)(T-T_{0}) \text{OPT}(\text{LP}(v^{*}))-\sum_{t=T_{0}+1}^{T_{1}} r\left(I_{t}\right)+\tau+\omega T \mathrm{OPT}\left(\mathrm{LP}\left(v^{*}\right)\right)\\
        	&\leq 4(\sqrt{N}+1)\sqrt{1+\frac{NT}{\tau q(L,K)}}\Psi\sqrt{N^{2}T}+\sqrt{2T\log\frac{4(K+1)}{\delta}}+\left(\frac{2N^{2} \Psi}{\sqrt{\tau}}+N\right) \sqrt{2T\log\frac{4(K+1)}{\delta}}+\tau+\\
        	&\frac{1}{\min _{k \in \mathcal{K}} c(k)}\left( \tau+4(\sqrt{N}+1)\sqrt{1+\frac{NT}{\tau q(L,K)}}\Psi\sqrt{N^{2}T}+\sqrt{2T\log\frac{4(K+1)}{\delta}}+\left(\frac{2N^{2} \Psi}{\sqrt{\tau}}+N \right)\sqrt{2T\log\frac{4(K+1)}{\delta}}\right)\\
        	&=(1+\frac{1}{\min _{k \in \mathcal{K}} c(k)})\left( \tau+4(\sqrt{N}+1)\sqrt{1+\frac{NT}{\tau q(L,K)}}\Psi\sqrt{N^{2}T}+\left(\frac{2N^{2} \Psi}{\sqrt{\tau}}+N+1 \right)\sqrt{2T\log\frac{4(K+1)}{\delta}}\right)
	    \end{split}
	\end{equation}

\section{Some Poofs of Models~\eqref{model:Choice_LP} and~\eqref{model:Compact_LP}}
\subsection{Proof of Theorem \ref{theorem:EquivLP}}

In practice, the Compact LP may have multiple optimal solutions. Let $z_{\mathrm{LP}}^{*}$ be the optimal objective value of the Compact LP, we can then solve another LP where we maximize $x_{0}$ in the objective function, subject to the constraint that $ \sum_{i \in N} r_{i}\Big[\left(\hat{v}_{i}^{\ell}+ \varepsilon(n^{\ell-1}_i)
\right) x_{i}+\varepsilon(n^{\ell-1}_i)\sum_{j \in N} \hat{v}_{j}^{\ell} y_{i j}\Big]\geq z_{L \mathrm{P}}^{*}$, along with all constraints in the Compact LP. In this way, we get an optimal solution with the largest value for the decision variable $x_{0}$, and an objective value of at least $z_{\mathrm{LP}}^{*}$. 

   To prove the Models~\eqref{model:Choice_LP} and ~\eqref{model:Compact_LP} are equivalent, we capture the polytope defined by these constraints as:
    $$\mathcal{P}=\left\{\left(x_{0}, \boldsymbol{x}, \boldsymbol{y}\right) \in \mathbb{R} \times \mathbb{R}_{+}^{n+n^{2}}:  x_{0}+\sum_{i \in N} \hat{v}_{i}^{\ell} x_{i}=1, x_{i} \leq x_{0}\quad\forall i \in N, y_{i j} \leq \min \left\{x_{i}, x_{j}\right\} \forall i, j \in N\right\} .$$ We construct the Lagrangian for the Compact LP by associating the dual multipliers $\boldsymbol{\mu}=\left\{\mu_{k
    }: k \in \mathcal{K}\right\}$ with the first constraint and relaxing this constraint, so the Lagrangian is
    \begin{equation}\label{equation:lagrangianRelax}
    \begin{aligned}
        L\left(x_{0}, \boldsymbol{x}, \boldsymbol{y} ; \boldsymbol{\mu}\right) &=\sum_{i \in N}  r_{i}\left(\left(\hat{v}_{i}^{\ell}+ \varepsilon(n^{\ell-1}_i)\right) x_{i}+\varepsilon(n^{\ell-1}_i) \sum_{j \in N} \hat{v}_{j}^{\ell} y_{i j}\right) \\
        &+\sum_{k \in \mathcal{K}} \mu_{k}\left((1-\omega)c_{k}-\sum_{i \in N}  a(i,k) \left(\left(\hat{v}_{i}^{\ell}-\varepsilon(n^{\ell-1}_i)\right) x_{i}-\varepsilon(n^{\ell-1}_i) \sum_{j \in N} \hat{v}_{j}^{\ell} y_{i j}\right)\right) \\
        &=\sum_{i \in N} r_{i}\left(\left(\hat{v}_{\ell}(i)+ \varepsilon(n^{\ell-1}_i)\right) x_{i}+\varepsilon(n^{\ell-1}_i) \sum_{j \in N} \hat{v}_{\ell}(j) y_{i j}\right)\\
        &-\sum_{k \in \mathcal{K}} a(i,k) \mu_{k}\left(\left(\hat{v}_{i}^{\ell}- \varepsilon(n^{\ell-1}_i)\right) x_{i}-\varepsilon(n^{\ell-1}_i) \sum_{j \in N} \hat{v}_{j}^{\ell} y_{i j}\right)+\sum_{k \in \mathcal{K}} (1-\omega)c_{k} \mu_{k} .
    \end{aligned}
    \end{equation}
    In this case, using $D(\boldsymbol{\mu})$ to denote the dual function for the Compact LP as a function of the dual multipliers $\boldsymbol{\mu}$, we have $D(\boldsymbol{\mu})=\max _{\left(x_{0}, \boldsymbol{x}, \boldsymbol{y}\right) \in \mathcal{P}} L\left(x_{0}, \boldsymbol{x}, \boldsymbol{y} ; \boldsymbol{\mu}\right)$. The Compact LP is feasible and bounded, so strong duality holds. Thus, we can obtain the optimal objective value of the Compact LP by solving the dual problem $\min _{\boldsymbol{\mu} \in \mathbb{R}_{+}^{m}} D(\boldsymbol{\mu})$.
    
    \begin{equation}\label{equation:D_mu}
    \begin{aligned}
        D(\boldsymbol{\mu})=& \max _{\left(x_{0}, \boldsymbol{x}, \boldsymbol{y}\right) \in \mathcal{P}}\left\{\sum_{i \in N} r_{i}\left(\left(\hat{v}_{i}^{\ell}+ \varepsilon(n^{\ell-1}_i)\right) x_{i}+\varepsilon(n^{\ell-1}_i) \sum_{j \in N} \hat{v}_{j}^{\ell} y_{i j}\right)\right.\\
        &\left.-\sum_{k \in \mathcal{K}} a(i,k) \mu_{k}\left(\left(\hat{v}_{i}^{\ell}- \varepsilon(n^{\ell-1}_i)\right) x_{i}-\varepsilon(n^{\ell-1}_i) \sum_{j \in N} \hat{v}_{j}^{\ell} y_{i j}\right)\right\}+\sum_{k \in \mathcal{K}} (1-\omega)c_{k} \mu_{k}  \\
        \stackrel{(a)}{=}& \max _{S \subseteq N} \left\{\sum_{i \in S} r_{i}\left(\frac{\hat{v}_{i}^{\ell}}{1+\sum_{j\in S}\hat{v}_{j}^{\ell}}+\varepsilon(n^{\ell-1}_i)\right)-\sum_{k \in \mathcal{K}} a(i,k) \mu_{k}\left(\frac{\hat{v}_{i}^{\ell}}{1+\sum_{j\in S}\hat{v}_{j}^{\ell}}-\varepsilon(n^{\ell-1}_i)\right)\right\}\\
        &+\sum_{k \in \mathcal{K}} (1-\omega)c_{k} \mu_{k} \\
        \stackrel{(b)}{=}& \max _{\boldsymbol{w} \in \mathbb{R}_{+}^{2}}\left\{\sum_{S \subseteq N} y_{\ell}(S)\sum_{i \in S} r_{i}\left(\frac{\hat{v}_{i}^{\ell}}{1+\sum_{j\in S}\hat{v}_{j}^{\ell}}+\varepsilon(n^{\ell-1}_i)\right)-\sum_{k \in \mathcal{K}} a(i,k) \mu_{k}\left(\frac{\hat{v}_{i}^{\ell}}{1+\sum_{j\in S}\hat{v}_{j}^{\ell}}-\varepsilon(n^{\ell-1}_i)\right):\right.\\
        &\left.\sum_{S \subseteq N} y_{l}(S)=1\right\}+\sum_{k \in \mathcal{K}} (1-\omega)c_{k} \mu_{k}  \\
        =& \max _{\boldsymbol{w} \in \mathbb{R}_{+}^{2^{n}}}\left\{\sum_{S \subseteq N} \sum_{i \in S}  r_{i}\left(\frac{\hat{v}_{i}^{\ell}}{1+\sum_{j\in S}\hat{v}_{j}^{\ell}}+\varepsilon(n^{\ell-1}_i)\right) y_\ell(S)\right.\\
        &\left.\quad+\sum_{k \in \mathcal{K}} \mu_{k}\left((1-\omega)c_{k}-\sum_{S \subseteq N} \sum_{i \in S} a(i,k)\left(\frac{\hat{v}_{i}^{\ell}}{1+\sum_{j\in S}\hat{v}_{j}^{\ell}}-\varepsilon(n^{\ell-1}_i)\right) y_\ell(S)\right): \sum_{S \subseteq N} y_\ell(S)=1\right\} .
    \end{aligned}
    \end{equation}
    In the chain of equations above, we first prove the equality $(a)$ holds, that is, prove the following problem (I) and (II) are equivalent.\\
    \textbf{Problem (I):}
    \begin{equation}
            \max _{S \subseteq N} \left\{\sum_{i \in S} r_{i}\left(\frac{\hat{v}_{i}^{\ell}}{1+\sum_{j\in S}\hat{v}_{j}^{\ell}}+\varepsilon(n^{\ell-1}_i)\right)-\sum_{k \in \mathcal{K}} a(i,k) \mu_{k}\left(\frac{\hat{v}_{i}^{\ell}}{1+\sum_{j\in S}\hat{v}_{j}^{\ell}}-\varepsilon(n^{\ell-1}_i)\right)\right\}
    \end{equation}
    \textbf{Problem (II):}
    \begin{equation}\label{eq54}
    \begin{split}
        \max _{\left(x_{0}, \boldsymbol{x}, \boldsymbol{y}\right) \in \mathbb{R} \times \mathbb{R}_{+}^{n+n^{2}}} & \sum_{i \in N} r_{i}\left(\left(\hat{v}_{i}^{\ell}+ \varepsilon(n^{\ell-1}_i)\right) x_{i}+\varepsilon(n^{\ell-1}_i) \sum_{j \in N} \hat{v}_{j}^{\ell} y_{i j}\right)\\
        &-\sum_{k \in \mathcal{K}} a(i,k) \mu_{k}\left(\left(\hat{v}_{i}^{\ell}- \varepsilon(n^{\ell-1}_i)\right) x_{i}-\varepsilon(n^{\ell-1}_i) \sum_{j \in N} \hat{v}_{j}^{\ell} y_{i j}\right)\\
       & x_{0}+\sum_{i \in N} \hat{v}_{j}^{\ell} x_{i}=1 \\
        & x_{i} \leq x_{0} \quad \forall i \in N \\
        &y_{i j} \leq x_{i} \quad \forall i, j \in N, \quad y_{i j} \leq x_{j} \quad \forall i, j \in N
    \end{split}
    \end{equation}
    \begin{proof}
       (1)  given a solution $\hat{S}\subseteq N$ to the Problem (I), we construct a solution ($x_{0}, \boldsymbol{x}, \boldsymbol{y}$) to the Problem (II) by setting $\hat{x}_{0}=\frac{1}{1+\sum_{i\in S}\hat{v}_{i}^{\ell}}, \quad \hat{x}_{i}=\mathbf{1}(i \in \hat{S}) \hat{x}_{0}$, and $\hat{y}_{i j}=\mathbf{1}(i \in \hat{S}, j \in \hat{S}) \hat{x}_{0}$. Noting that $\sum_{i \in N} \hat{v}_{i}^{\ell} \hat{x}_{i}=\hat{x}_{0} \sum_{i \in N} \hat{v}_{i}^{\ell} \mathbf{1}(i \in \hat{S})=\hat{x}_{0} \sum_{i\in S}\hat{v}_{i}^{\ell}$, we have $\hat{x}_{0}+\sum_{i \in N} \hat{v}_{i}^{\ell} \hat{x}_{i}=\hat{x}_{0}(1+\sum_{i\in S}\hat{v}_{i}^{\ell})=1$, so the solution ($x_{0}, \boldsymbol{x}, \boldsymbol{y}$) satisfies the first constraint in the Problem (II). Moreover, since $\mathbf{1}(i \in \hat{S}) \leq 1$, $\mathbf{1}(i \in \hat{S}, j \in \hat{S}) \leq \mathbf{1}(i \in \hat{S})$, and $\mathbf{1}(i \in \hat{S}, j \in \hat{S}) \leq \mathbf{1}(j \in \hat{S})$, the solution $\left(\hat{x}_{0}, \hat{\boldsymbol{x}}, \hat{\boldsymbol{y}}\right)$ satisfies the remaining constraints in the Problem (II) as well. Furthermore, for the Problem (II), this solution provides an objective value of
    \begin{equation}\small
        \begin{split}
                &\sum_{i \in N} r_{i}\left(\left(\hat{v}_{i}^{\ell}+ \varepsilon(n^{\ell-1}_i)\right) x_{i}+\varepsilon(n^{\ell-1}_i) \sum_{j \in N} \hat{v}_{\ell}(j) y_{i j}\right)-\sum_{k \in \mathcal{K}} a(i,k) \mu_{k}\left(\left(\hat{v}_{i}^{\ell}- \varepsilon(n^{\ell-1}_i)\right) x_{i}-\varepsilon(n^{\ell-1}_i) \sum_{j \in N} \hat{v}_{\ell}(j) y_{i j}\right)\\
                =&\sum_{i \in N} r_{i}\left(\left(\hat{v}_{i}^{\ell}+ \varepsilon(n^{\ell-1}_i)\right) +\varepsilon(n^{\ell-1}_i) \sum_{j \in N} \hat{v}_{j}^{\ell} \mathbf{1}( j \in \hat{S})\right)\hat{x}_{0}\mathbf{1}(i \in \hat{S})\\
                &-\sum_{k \in \mathcal{K}} a(i,k) \mu_{k}\left(\left(\hat{v}_{i}^{\ell}- \varepsilon(n^{\ell-1}_i)\right) -\varepsilon(n^{\ell-1}_i) \sum_{j \in N} \hat{v}_{j}^{\ell}\mathbf{1}(j \in \hat{S}) \right)\hat{x}_{0}\mathbf{1}(i \in \hat{S})\\
                =&\sum_{i \in N} r_{i}\left(\hat{v}_{i}^{\ell}+ \varepsilon^{\ell-1}_i\left(1+\sum_{j \in \hat{S}} \hat{v}_{j}^{\ell} \right)\right)\hat{x}_{0}\mathbf{1}(i \in \hat{S})-\sum_{k \in \mathcal{K}} a(i,k) \mu_{k}\left(\hat{v}_{i}^{\ell}-\varepsilon^{\ell-1}_i\left(1+\sum_{j \in \hat{S}} \hat{v}_{j}^{\ell} \right)\right)\hat{x}_{0}\mathbf{1}(i \in \hat{S})\\
                =&\sum_{i \in \hat{S}} r_{i}\left(\frac{\hat{v}_{i}^{\ell}}{1+\sum_{j\in \hat{S}}\hat{v}_{j}^{\ell}}+\varepsilon(n^{\ell-1}_i)\right)-\sum_{k \in \mathcal{K}} a(i,k) \mu_{k}\left(\frac{\hat{v}_{i}^{\ell}}{1+\sum_{i\in \hat{S}}\hat{v}_{j}^{\ell}}-\varepsilon(n^{\ell-1}_i)\right)
        \end{split}
    \end{equation}
which is the objective function of the Problem (I) evaluated at $\hat{S}$. Thus, given a solution to
the  Problem (I), we can construct a feasible solution to the  Problem (II) , and the objective values of the two solutions match. Now, we prove the converse statement as well.\\
(2) To establish the converse statement, we build on the next lemma, which shows an important property of the basic feasible solutions to the Problem (II).

\noindent{\textbf{Lemma C.1 (Extreme Point Solutions).}} Let $\left(\hat{x}_{0}, \hat{\boldsymbol{x}}, \hat{\boldsymbol{y}}\right)$ be a basic feasible solution to the Problem (II). Then, we have $\hat{x}_{i} \in\left\{0, \hat{x}_{0}\right\}$ for all $i \in N$.

\noindent{\textbf{Theorem $C.1$ }} For a basic optimal solution $\left(x_{0}^{*}, \boldsymbol{x}^{*}, \boldsymbol{y}^{*}\right)$ to the Problem (II), let $S^{*}=\left\{i \in N: x_{i}^{*}\neq0\right\}$. Then, $S^{*}$ is an optimal solution to the Problem (I).\\
Proof: Let $\hat{S}$ be an optimal solution to the Problem (I) providing the optimal objective value $\hat{z}$, and let $z_{\mathrm{LP}}^{*}$ be the optimal objective value of the Problem (II) . By the discussion above, $z_{\mathrm{LP}}^{*}\geq \hat{z}$. On the other hand, by Lemma C.1, we have $x_{i}^{*}=x_{0}^{*}$ for all $i \in S^{*}$ and $x_{i}^{*}=0$ for all $i \in N \backslash S^{*}$. Since $\left(x_{0}^{*}, \boldsymbol{x}^{*}, \boldsymbol{y}^{*}\right)$ is a feasible solution to the Assortment LP, by the first constraint, we get $x_{0}^{*}+\sum_{i \in S^{*}} \hat{v}_{i}^{\ell} x_{0}^{*}=1$, so $x_{0}^{*}=\frac{1}{1+\sum_{i\in S^{*}} \hat{v}_{i}^{\ell}}=x_{i}^{*}$ for all $i \in S^{*}$. In this case, by the last two constraints, we also get $y_{i j}^{*} \leq \frac{1}{1+\sum_{i\in S^{*}} \hat{v}_{i}^{\ell}}$ for all $i, j \in S^{*}$. If $i \notin S^{*}$ or $j \notin S^{*}$, then $x_{i}^{*}=0$ or $x_{j}^{*}=0$, so we have $y_{i j}^{*}=0$. 
\begin{equation}\label{equation:hat_z}
\begin{aligned}
    z_{\mathrm{LP}}^{*} &=\sum_{i \in S^{*}} r_{i}\left(\left(\hat{v}_{i}^{\ell}+ \varepsilon(n^{\ell-1}_i)\right) x_{i}^{*}+\varepsilon(n^{\ell-1}_i) \sum_{j \in S^{*}} \hat{v}_{j}^{\ell} y_{i j}^{*}\right)\\
    &-\sum_{k \in \mathcal{K}} a(i,k) \mu_{k}\left(\left(\hat{v}_{i}^{\ell}- \varepsilon(n^{\ell-1}_i)\right) x_{i}^{*}-\varepsilon(n^{\ell-1}_i) \sum_{j \in S^{*}} \hat{v}_{j}^{\ell} y_{i j}^{*}\right)\\
    & {\leq} \sum_{i \in S^{*}} r_{i}\left(\frac{\hat{v}_{i}^{\ell}+ \varepsilon(n^{\ell-1}_i)}{1+\sum_{j\in S^{*}}\hat{v}_{j}^{\ell}} +\frac{\varepsilon(n^{\ell-1}_i)}{1+\sum_{j\in S^{*}}\hat{v}_{j}^{\ell}} \sum_{j \in S^{*}} \hat{v}_{j}^{\ell}\right)\\
    &-\sum_{k \in \mathcal{K}} a(i,k) \mu_{k}\left(\frac{\hat{v}_{i}^{\ell}- \varepsilon(n^{\ell-1}_i)}{1+\sum_{j\in S^{*}}\hat{v}_{j}^{\ell}} -\frac{\varepsilon(n^{\ell-1}_i)}{1+\sum_{j\in S^{*}}\hat{v}_{j}^{\ell}} \sum_{j \in S^{*}} \hat{v}_{j}^{\ell}\right)\\
    &=\sum_{i \in S^{*}} r_{i}\left(\frac{\hat{v}_{i}^{\ell}}{1+\sum_{j\in S^{*}}\hat{v}_{j}^{\ell}} +\varepsilon(n^{\ell-1}_i)\right)-\sum_{k \in \mathcal{K}} a(i,k) \mu_{k}\left(\frac{\hat{v}_{i}^{\ell}}{1+\sum_{j\in S^{*}}\hat{v}_{j}^{\ell}} +\varepsilon(n^{\ell-1}_i)\right) {\leq} \hat{z}
\end{aligned}
\end{equation}
Combing the two parts (1) and (2), we conclude that $z_{\mathrm{LP}}^{*}=\hat{z}$, which further implies the problems (I) and (II) are equivalent.
    \end{proof}
 Since picking one assortment that maximizes the expected revenue is equivalent to randomizing over all possible assortments, the inequality (b) in the chain of  Eq.(\ref{equation:D_mu}) holds.  In this case, if we minimize the dual functions for the two LP formulations over all $\boldsymbol{\mu} \in \mathbb{R}_{+}^{m}$, then we get the same optimal objective value, so the two LP formulations have the same optimal objective value. Furthermore, the minimizers of the dual functions for the two LP formulations must be the same, which implies that the optimal values of the dual variables for the first constraint in the two LP formulations are the same.

\subsection{ Proof of Theorem ~\ref{theorem:RecoverOPT}}
For notational brevity, let $\Lambda_{i}^{1}=\left(\hat{v}_{i}^{\ell}+ \varepsilon(n^{\ell-1}_i)\right) x_{i}^{*}+\varepsilon(n^{\ell-1}_i) \sum_{j \in S^{*}} \hat{v}_{j}^{\ell} y_{i j}^{*}$, $\Lambda_{i}^{2}=\left(\hat{v}_{i}^{\ell}- \varepsilon(n^{\ell-1}_i)\right) x_{i}^{*}-\varepsilon(n^{\ell-1}_i) \sum_{j \in S^{*}} \hat{v}_{j}^{\ell} y_{i j}^{*}$. By Theorem \ref{theorem:EquivLP}, the optimal objective values of the Compact LP and UCB-LP models are equal. Let $z_{\mathrm{LP}}^{*}$ be their common optimal objective value. Noting the objective function of the Compact LP, we have $z_{\mathrm{LP}}^{*}=\sum_{i \in N} r_{i} \Lambda_{i}^{1}$. Furthermore, by the first constraint in the Compact $\mathrm{LP}$, we have $\sum_{i \in N} a(i,k) \Lambda_{i}^{2} \leq c_{k}$. We can derive that $\sum_{S \subseteq N} \hat{w}(S)=1$ and
\begin{equation}
   \sum_{S \subseteq N} \mathbf{1}(i \in S)\left(\frac{\hat{v}_{i}^{\ell}}{1+\sum_{j\in S}\hat{v}_{j}^{\ell}} +\varepsilon(n^{\ell-1}_i)\right) \hat{w}(S)=\Lambda_{i}^{1} 
\end{equation}
\begin{equation}
  \sum_{S \subseteq N} \mathbf{1}(i \in S)\left(\frac{\hat{v}_{i}^{\ell}}{1+\sum_{j\in S}\hat{v}_{j}^{\ell}} -\varepsilon(n^{\ell-1}_i)\right) \hat{w}(S)=\Lambda_{i}^{2}  
\end{equation}
In this case, we further get
\begin{equation}
\sum_{S \subseteq N} \sum_{i \in S} r_{i}\left(\frac{\hat{v}_{i}^{\ell}}{1+\sum_{j\in S}\hat{v}_{j}^{\ell}} +\varepsilon(n^{\ell-1}_i)\right) \hat{w}(S)=\sum_{i \in N} r_{i} \Lambda_{i}^{1}=z_{\mathrm{Lp}}^{*} 
\end{equation}
 \begin{equation}
     \sum_{S \in \mathcal{N}}\sum_{i \in S}a(i,k)\left(\frac{\hat{v}_{i}^{\ell}}{1+\sum_{j\in S}\hat{v}_{j}^{\ell}}-\varepsilon(n^{\ell-1}_i)\right) \hat{w}(S)\leq(1-\omega) c(k)
 \end{equation}
Thus, the solution $\hat{\boldsymbol{w}}$ provides an objective value of $z_{\mathrm{LP}}^{*}$ for the UB-LP model.

\end{document}